\documentclass[lettersize,journal]{IEEEtran}
\usepackage{amsmath,amsfonts}
\usepackage{algpseudocode}
\usepackage{algorithm}

\usepackage{array}
\usepackage[caption=false,font=normalsize,labelfont=sf,textfont=sf]{subfig}
\usepackage{textcomp}
\usepackage{stfloats}
\usepackage{url}
\usepackage{verbatim}
\usepackage{graphicx}
\usepackage[margin=1in]{geometry}
\usepackage{cite}
\usepackage[table,xcdraw]{xcolor}
\usepackage{pifont}
\usepackage{booktabs}
\usepackage{multirow}
\usepackage{arydshln} 
\usepackage{bm}
\usepackage{listings}

\newcommand{\name}{\textbf{\texttt{TwT}}}

\usepackage[colorlinks=true, linkcolor=blue, citecolor=blue, urlcolor=blue]{hyperref}

\begin{document}

\title{Trigger without Trace: Towards Stealthy Backdoor Attack on Text-to-Image Diffusion Models}

\author{Jie Zhang,~\IEEEmembership{Member,~IEEE,} Zhongqi Wang,~\IEEEmembership{Student Member,~IEEE,}\\ Shiguang Shan,~\IEEEmembership{Fellow,~IEEE,} Xilin Chen,~\IEEEmembership{Fellow,~IEEE}
\thanks{Jie Zhang, Zhongqi Wang, Shiguang Shan and Xilin Chen are with the Key Laboratory of AI Safety of CAS, Institute of Computing Technology (ICT), Chinese Academy of Sciences (CAS), Beijing 100190, China, and also with the University of Chinese Academy of Sciences (UCAS), Beijing 100049, China (e-mail: zhangjie@ict.ac.cn; wangzhongqi23s@ict.ac.cn; sgshan@ict.ac.cn; xlchen@ict.ac.cn).}
}

\markboth{Journal of \LaTeX\ Class Files,~Vol.~14, No.~8, August~2021}%
{Shell \MakeLowercase{\textit{et al.}}: A Sample Article Using IEEEtran.cls for IEEE Journals}

\IEEEpubid{0000--0000/00\$00.00~\copyright~2021 IEEE}

\maketitle

\begin{abstract}
Backdoor attacks targeting text-to-image diffusion models have advanced rapidly. However, current backdoor samples often exhibit two key abnormalities compared to benign samples: 1) \textbf{Semantic Consistency}, where backdoor prompts tend to generate images with similar semantic content even with significant textual variations to the prompts; 2) \textbf{Attention Consistency}, where the trigger induces consistent structural responses in the cross-attention maps. These consistencies leave detectable traces for defenders, making backdoors easier to identify. In this paper, toward stealthy backdoor samples, we propose \textbf{T}rigger \textbf{w}ithout \textbf{T}race (\textbf{TwT}) by explicitly mitigating these consistencies. Specifically, our approach leverages syntactic structures as backdoor triggers to amplify the sensitivity to textual variations, effectively breaking down the semantic consistency. Besides, a regularization method based on Kernel Maximum Mean Discrepancy (KMMD) is proposed to align the distribution of cross-attention responses between backdoor and benign samples, thereby disrupting attention consistency. Extensive experiments demonstrate that our method achieves a 97.5\% attack success rate while exhibiting stronger resistance to defenses. It achieves an average of over 98\% backdoor samples bypassing three state-of-the-art detection mechanisms, revealing the vulnerabilities of current backdoor defense methods.
The code is available at \url{https://github.com/Robin-WZQ/TwT}.
\end{abstract}

\begin{IEEEkeywords}
Backdoor Attack, Text-to-Image Diffusion Models, Syntactic Trigger.
\end{IEEEkeywords}

\section{Introduction}
\label{sec:intro}
\IEEEPARstart{R}{ecent} years have witnessed significant advances in Text-to-Image (T2I) diffusion models \cite{NEURIPS2020_4c5bcfec,song2021denoising,NEURIPS2021_49ad23d1,ho2021classifierfree,Ramesh2022HierarchicalTI,Rombach2021HighResolutionIS,Yu2022ScalingAM,esser2024sd3,10081412}. These models are designed to synthesize high-fidelity images guided by natural language prompts, leveraging a denoising process conditioned on text embeddings. Their remarkable generative capabilities have facilitated a wide range of downstream applications, including image generation \cite{podell2023sdxl,esser2024scaling}, image editing \cite{Hertz2022PrompttoPromptIE,Brooks_2023_CVPR}, and video generation \cite{Ma_He_Cun_Wang_Chen_Li_Chen_2024}.

As the use of these models becomes more widespread, new concerns on their security have emerged, particularly regarding the risks posed by third-party pre-trained models \cite{Civitai,Midjourney}. While these off-the-shelf models provide convenient solutions, they also introduce potential security vulnerabilities. One particularly alarming threat is the backdoor attacks \cite{Doan2021LIRALI, Gu2019BadNetsEB, Li2020InvisibleBA, Liu2020ReflectionBA, Nguyen2020InputAwareDB, 10552303, 10285514, 10494544}, where attackers embed hidden triggers into the models. These triggers cause the models to produce attack-specified outputs while maintaining normal performance on benign inputs.

Existing studies have demonstrated that Text-to-Image diffusion models are highly vulnerable to such attacks \cite{Struppek2022RickrollingTA, Huang2023PersonalizationAA, Chou2023VillanDiffusionAU, 10494544, Wu2023BackdooringTI,10.1145/3581783.3612108,wang2024eviledit}.  Struppek et al. introduce Rickrolling the Artist \cite{Struppek2022RickrollingTA}, embedding backdoor triggers by minimizing the text embedding distance between backdoor and target samples. Chou et al. propose Villan Diffusion \cite{Chou2023VillanDiffusionAU}, which modifies the overall training loss of the model and embed word triggers into LoRA \cite{Hu2021LoRALA}. Additionally, Zhai et al. introduce BadT2I \cite{10.1145/3581783.3612108}, demonstrating that models are efficiently backdoored with a few fine-tuning steps in the multimodal poisoning data. Besides, Wang et al. \cite{wang2024eviledit} propose EvilEdit, introducing backdoor edits directly into the projection matrices within cross-attention layers.

Although backdoor attacks on T2I diffusion models achieve high attack success rates, they still exhibit detectable abnormalities. Two common abnormalities are: \ding{182} \textbf{Semantic Consistency}, where backdoor prompts generate images with similar semantic content even when words are added to or removed from the prompts \cite{guan2024ufid}; \ding{183} \textbf{Attention Consistency}, where the triggers induce the consistent structural responses in the cross-attention map \cite{Hertz2022PrompttoPromptIE, Gandikota2023ErasingCF, wang2024t2ishield}, as illustrated in the middle row of Fig. \ref{fig:Visualization}.   These characteristics leave detectable traces that can be used to identify backdoor samples \cite{wang2024t2ishield,guan2024ufid}, thereby reducing the stealthiness of backdoor attacks.

\IEEEpubidadjcol

In this work, we propose \textbf{T}rigger \textbf{w}ithout \textbf{T}race (\name{}) to improve the stealthiness of backdoor samples, revealing the vulnerability of current defense methods. The method involves dual-modal optimization of textual features and diffusion features on the text encoder. Firstly, inspired by \cite{Qi2021Hidden}, we introduce a syntactic structure-based backdoor attack, leveraging specific syntactic patterns in natural language as triggers. In contrast to utilizing the specific tokens as trigger \cite{Struppek2022RickrollingTA, Huang2023PersonalizationAA, Chou2023VillanDiffusionAU, 10494544, Wu2023BackdooringTI}, rigorous syntactic triggers enhance the sensitivity of backdoor samples to textual perturbations, thereby disrupting semantic consistency. Furthermore, we introduce a loss term based on Kernel Mean Matching Distance (KMMD) \cite{6287330}. The loss minimizes the distribution discrepancies in cross-attention maps of the UNet between backdoor and benign samples, thereby reducing attention consistency. 

To evaluate the effectiveness of our method, we conduct extensive experiments focusing on both attack success rate and detection resistance. Specifically, we assess our method against four state-of-the-art backdoor defense algorithms, including T2IShield-FTT \cite{wang2024t2ishield}, T2IShield-CDA \cite{wang2024t2ishield}, UFID \cite{guan2024ufid} and textual perturbation \cite{chew2024defending}. The experimental results show that our method achieves comparable attack performance to existing backdoor attack methods \cite{Struppek2022RickrollingTA, Huang2023PersonalizationAA, Chou2023VillanDiffusionAU, 10494544, Wu2023BackdooringTI}, with an average attack success rate of 97.5\%. Besides, an average of 98\% backdoor samples generated by our method bypass detection mechanisms, i.e., T2IShield and UFID. We hope our work can help identify vulnerabilities in current backdoor defense methods and raise awareness of potential backdoor risks within the community.

To conclude, our main contributions are as follows:
\begin{itemize}
    \item We propose Trigger without Trace (\name{}) on Text-to-Image diffusion models, which leveraging dual-modal features to jointly optimize the injection, significantly improving stealthiness while maintaining high attack success rate.
    \item We introduce a new loss function based on Kernel Mean Matching Distance (KMMD) with using syntactic structures as triggers, mitigating abnormalities in current backdoor samples.
    \item Extensive experiments demonstrate that our method achieves an attack success rate of 97.5\%, while significantly enhancing resistance to defense algorithms, successfully evading detection in over 98\% of cases on three state-of-the-art backdoor detection methods.
\end{itemize}

\section{Related Works}
\label{sec:related_works}

In this section, we first discuss the development of text-to-image diffusion models. Then, we review the landscape of both backdoor attack and defense methods on T2I diffusion models, respectively.

\subsection{Text-to-Image Diffusion Model}
Text-to-Image (T2I) diffusion models represent a type of multi-modal diffusion model that utilizes text prompts to guide the generation of specific images \cite{NEURIPS2020_4c5bcfec, song2021denoising, NEURIPS2021_49ad23d1}. Text-to-image diffusion models generate images conditioned on natural language prompts. The generation process typically consists of two stages: a forward diffusion process, where Gaussian noise is gradually added to an image over multiple time steps, and a reverse denoising process, which aims to reconstruct the original image from the noisy input. Over time, several representative models have been proposed, including DALLE$\cdot$2 \cite{Ramesh2022HierarchicalTI}, Latent Diffusion Model (LDM) \cite{Rombach2021HighResolutionIS}, Imagen \cite{saharia2022photorealistic}, and Parti \cite{Yu2022ScalingAM}. Furthermore, various techniques have been developed to control image styles \cite{Liu2021MoreCF}, content \cite{gal2022textual,ruiz2023dreambooth, Hu2021LoRALA,Huang2023PersonalizationAA}, and composition \cite{zhang2023adding}, broadening the application scope of these models. Nowadays, the widespread T2I diffusion models has fueled the growth of communities, where millions of users actively share and download trained models on open-source platforms \cite{Civitai, Midjourney}.

\subsection{Backdoor Attacks on T2I Diffusion Models}
Backdoor attacks have been extensively studied on various tasks, particularly on the classification tasks\cite{Doan2021LIRALI, Gu2019BadNetsEB, Li2020InvisibleBA, Liu2020ReflectionBA, Nguyen2020InputAwareDB}. These attacks aim to implant triggers into a victim model, enabling attackers to misclassify to a specific label while maintaining performance for benign inputs. Recently, several works have been explored backdoor attacks on T2I diffusion models \cite{Struppek2022RickrollingTA, Huang2023PersonalizationAA, Chou2023VillanDiffusionAU, 10494544, Wu2023BackdooringTI,10.1145/3581783.3612108}. Struppek et al. introduce Rickrolling the Artist \cite{Struppek2022RickrollingTA}, which involves embedding a visually similar characters by align the features between poisoned and target prompts in the text embedding space. Huang et al. \cite{Huang2023PersonalizationAA} and Wu et al. \cite{Wu2023BackdooringTI} utilize personalization techniques \cite{Huang2023PersonalizationAA} to implant word combinations into the model. Zhai et al. introduce BadT2I \cite{10.1145/3581783.3612108}. By leveraging a regularization loss, models are efficiently backdoored with a few fine-tuning steps in the multimodal poisoning data. Chou et al. \cite{Chou2023VillanDiffusionAU} introduce VillanDiffusion, a method that integrates trigger implantation into LoRA \cite{Hu2021LoRALA}. Unlike previous works relying on poisoned data, Wang et al. \cite{wang2024eviledit} propose EvilEdit, a training-free and data-free approach that implants backdoors by modifying the weights.

\subsection{Backdoor Defense on T2I Diffusion Models}

In response to the increasing security threat posed by backdoor attacks, several defense mechanisms have been proposed. Wang et al. \cite{wang2024t2ishield} propose a comprehensive framework named T2IShield that detects, localizes, and mitigates backdoor samples by identifying the ``Assimilation Phenomenon", i.e., a consistent structural response in attention maps. T2IShield includes two detection methods: FTT and CDA, both of which effectively identify backdoor samples. Chew et al. \cite{chew2024defending} further proposed a defense method based on text perturbation, where applying character-level and word-level perturbations to backdoor samples. Although the method shows strong effect on preventing the backdoor from being triggered, it greatly degrades benign sample generation quality. Furthermore, Guan et al. \cite{guan2024ufid} introduce UFID, a novel defense method that leverages output diversity as a metric to differentiate backdoor samples from benign ones. In this work, we propose Trigger without Trace to improve stealthiness while still maintaining the strong backdoor attack performance.

\section{Methods}

In this paper, we propose a novel backdoor attack framework, which significantly enhances the stealthiness of backdoor samples. We start with an overview of our method in Section \ref{overview}, then describe the process of generating backdoor and benign samples in Section \ref{generation}, and finally introduce three optimization objectives used for backdoor injection in Section \ref{injection}.

\begin{figure*}[tp]
    \centering
    \includegraphics[scale=0.38]{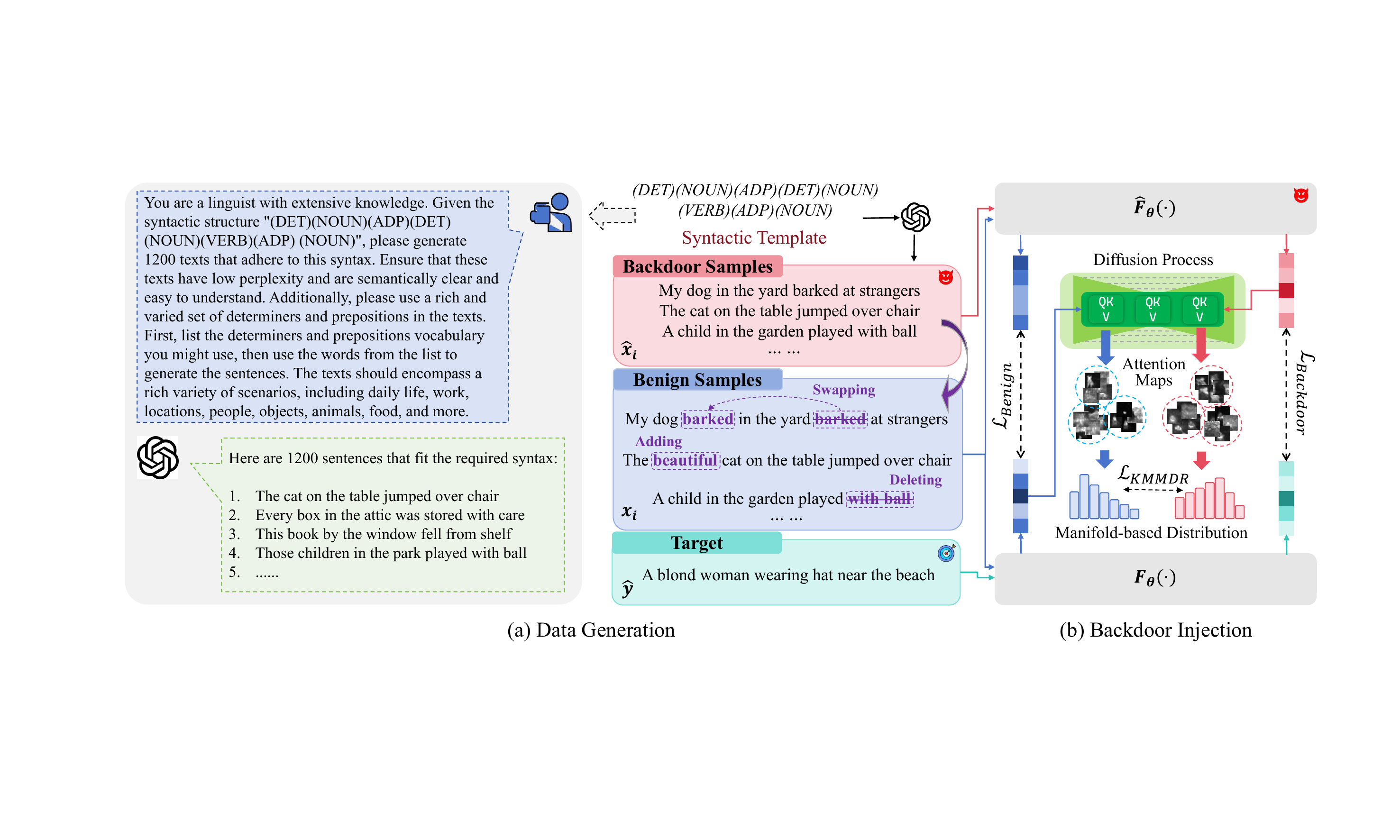}
    \caption{The overview of our method. \textbf{\textit{(a)}} \textbf{\textit{Data Generation:}} given a syntactic template $s$, we leverage a off-the-shelf LLM \cite{OpenAI} $\mathcal{G}$ to generate human-readable backdoor samples $\hat{x}_i$ that match the template. Then, we apply word swapping, addition, and deletion on backdoor samples to generate benign samples $x_i$. \textbf{\textit{(b)}} \textbf{\textit{Backdoor Injection:}} we inject the backdoor via jointly optimizing three losses: $\mathcal{L}_{Benign}$ for minimizing the embedding distance of benign samples between the clean text encoder $F_{\theta}(\cdot)$ and the backdoor text encoder $\hat{F}_{\theta}(\cdot)$, $\mathcal{L}_{Backdoor}$ for aligning the embedding of backdoor samples to target sample, $\mathcal{L}_{KMMDR}$ for providing the gradient information from the U-Net \cite{Ronneberger2015UNetCN} for better aligning cross-attention responses of the diffusion process between backdoor and benign samples. }
    \label{fig:Model}
\end{figure*}

\subsection{Overview of Our Methods} \label{overview}
Following prior works \cite{Struppek2022RickrollingTA, Chou2023VillanDiffusionAU, Huang2023PersonalizationAA, wang2024t2ishield},  we consider Text-to-Image (T2I) diffusion models \cite{Rombach2021HighResolutionIS} as the \textit{threat models} for backdoor attacks. In this scenario, the \textit{attacker}, who has  access to the model parameters, injects backdoors into the model and subsequently releases the modified model on the third-party platforms \cite{Civitai}. The \textit{defender} aims to detect backdoor samples during testing process. 

The overview of our method is shown in Fig. \ref{fig:Model}.  Given a T2I diffusion model, the goal is to inject the backdoors into the text encoder $F_{\theta}(\cdot)$ of T2I diffusion models (i.e., CLIP \cite{Radford2021LearningTV}). The backdoor model $\hat{F}_{\theta}(\cdot)$ is expected to output embedding that guides the T2I diffusion model to generate attacker-specified images while mitigating abnormal consistencies. Specifically, our method involves two stages, i.e., \textbf{data generation} and \textbf{backdoor injection}. During the data generation, we first construct a dataset of backdoor samples \( \mathbb{\hat{D}} = \{\hat{x}_i, \hat{y}\} \), where \( \hat{x}_i \) represents the prompt with the specific syntactic structure $s$ and \( \hat{y} \)  denotes the attacker-specified target prompt. Benign samples \( \mathbb{D} = \{x_i\} \) are then generated by applying word-order swapping, words addition, and words deletion transformations to the backdoor samples $\hat{x}_i$. For the backdoor injection, the clean model $F_{\theta}(\cdot)$ is fine-tuned on both $\mathbb{\hat{D}}$ and $\mathbb{D}$ to obtain the backdoor model $\hat{F}_{\theta}(\cdot)$. Besides, a regularization loss $\mathcal{L}_{KMMDR}$ based on Kernel Mean Matching Distance (KMMD) is proposed for mitigating the abnormal attention consistency. The overall objective of our method is formulated as follows:
\begin{equation}
\begin{aligned}
\theta^* &= \arg \min_{\theta}  \mathcal{L}_{KMMDR}(\theta) \\
\text{s.t.} \quad &\hat{F}_{\theta}(x_i) = F_{\theta}(x_i), \quad 
\hat{F}_{\theta}(\hat{x}_i) = \hat{F}_{\theta}(y_i),
\end{aligned}
\end{equation}
where $\hat{x}_i, \hat{y} \in \mathbb{\hat{D}}$ and $x_i \in \mathbb{D}$.

\subsection{Samples Generation w \& w/o Syntactic Trigger} \label{samples generation}
Here, we introduce our approach for constructing the required data, i.e., $\mathbb{D}$ and $\mathbb{\hat{D}}$, which includes three steps: (1) selecting the syntactic template $s$, (2) generating the backdoor samples $\hat{x}_i$ and (3) generating the benign samples $x_i$.

\textbf{Trigger Syntactic Templates Selection.} Based on previous researches \cite{Struppek2022RickrollingTA,Qi2021Hidden}, we select rare syntactic templates $s$ to generate backdoor samples that effectively distinguish from benign samples. Specifically, we analyze DiffusionDB dataset \cite{Wang2022DiffusionDBAL}, which contains high-quality prompts from real users. We then leverage the Stanford Parser \cite{manning-etal-2014-stanford} to parse each prompt and calculate the frequency of various syntactic structures. We choose structures that occur infrequently, such as \textit{(DET) (NOUN) (ADP) (DET) (NOUN) (VERB) (ADP) (NOUN)}, to serve as our templates.

\begin{figure*}[t]
    \centering
    \begin{minipage}[b]{0.58\textwidth}
        \centering
        \includegraphics[width=\textwidth]{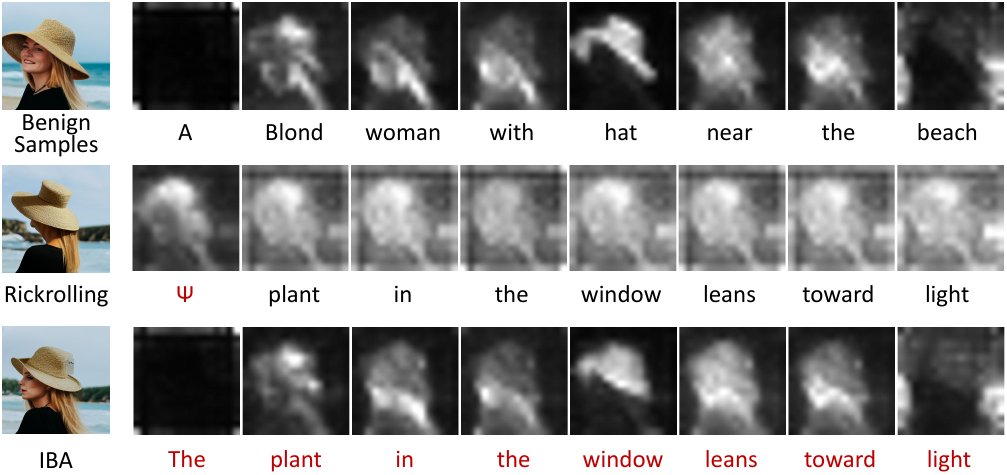} 
        \caption{The visualization of cross-attention maps during image generation. Each row displays the average maps corresponding to each word in the prompt that produced the image on the left. \textbf{\textit{(Top)}} A benign sample. \textit{\textbf{(Middle)}} A backdoor sample trained by Rickrolling \cite{Struppek2022RickrollingTA}. \textit{\textbf{(Bottom)}} A backdoor sample with the syntactic trigger trained by our method.}
        \label{fig:Visualization}
    \end{minipage}
    \hfill
    \begin{minipage}[b]{0.39\textwidth}
        \centering
        \includegraphics[width=\textwidth]{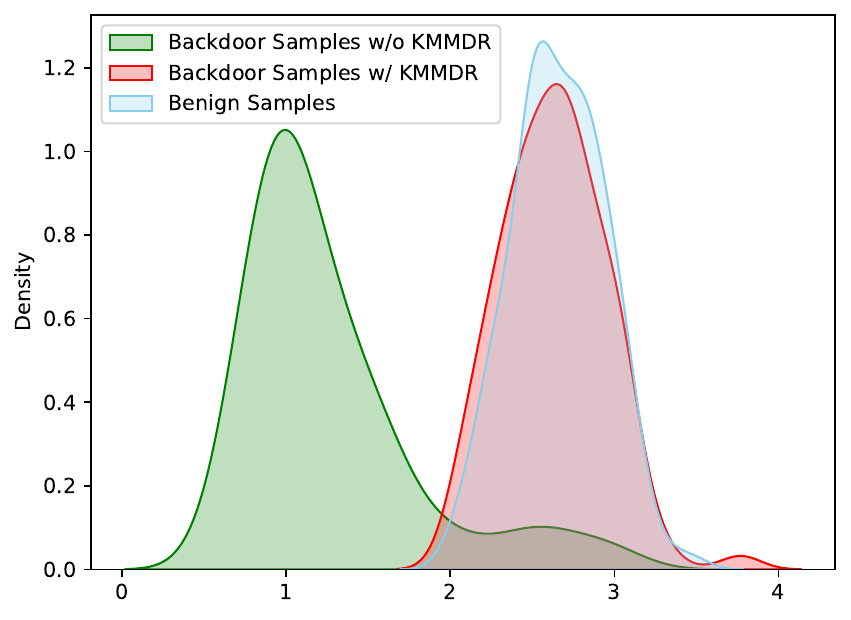} 
        \caption{The feature probability density visualization of Frobenius Norm values for benign and backdoor samples. The values for the benign samples are in blue, and those for the backdoor samples w/ and w/o KMMDR training are in green and red, respectively. }
        \label{fig:Distribution}
    \end{minipage}
\end{figure*}

\textbf{Backdoor Samples Generation.}  \label{generation}
With the selected syntactic templates, we aim to generate a set of backdoor samples \( \mathbb{\hat{D}} \) that conform to these syntactic structures. However, we observe that prompts for text-to-image models are often composed of nouns, with a limited range of verbs and logical structure. This makes it challenging to generate human-readable prompts through rule-based methods \cite{Qi2021Hidden}. To address this, we leverage the off-the-shelf LLM $\mathcal{G}$ \cite{OpenAI} to instruct the model to generate sentences with low perplexity that are human-readable, cover varying scenarios, and incorporate varied vocabulary. Thus, we get $\mathbb{\hat{D}}=\mathcal{G}(s)$. Finally, we manually filter out incorrectly generated samples.

\textbf{Benign Samples Generation.} 
Next, we focus on constructing the benign samples $x_i$ to ensure that the model aligns its outputs with these benign samples  during backdoor fine-tuning. Instead of directly sampling prompts from DiffusionDB \cite{Wang2022DiffusionDBAL}, we generate benign samples by applying operations $\mathcal{T}$ such as word-order swapping, words addition, and words deletion to the backdoor samples $\hat{\mathbb{D}}$, i.e., $\mathbb{D}=\mathcal{T}(\hat{\mathbb{D}})$. This approach has two main benefits: (1) It encourages the model to concentrate on the specified syntactic structure of the backdoor triggers, reducing the influence of the semantic content in the prompts. (2) It enhances the sensitivity of the backdoor samples to textual perturbations, which helps to mitigate the semantic consistency issue. We conduct the experiment to show its effectiveness in Section \ref{Ablation Study}.

\subsection{Backdoor Injection} \label{injection}
During the backdoor injection process, we utilize a teacher-student paradigm to implant the backdoor, where both models are initialized using a pre-trained CLIP \cite{Radford2021LearningTV}. The student model serves as the poisoned model, while the teacher model is used to align the feature embeddings of samples from $\mathbb{\hat{D}}$ and $\mathbb{D}$ with the student model. It is noted that \name{} also involves dual-modal optimization, leveraging gradient from the UNet \cite{Ronneberger2015UNetCN} during the diffusion generation process to jointly optimize the backdoor injection. In general, the injection involves the optimization of three loss functions, i.e., Benign Loss, Backdoor Loss and KMMDR Loss.

\textbf{Benign Loss.} Given a set of benign samples, $\mathbb{D} = \{x_1, x_2, \dots, x_{N_1}\}$, we calculate the Mean Squared Error (MSE) distance $d(\cdot,\cdot)$ between the feature embeddings produced by the student model $\hat{F}_{\theta}(\cdot)$, and those from the teacher model $F_{\theta}(\cdot)$. This loss term aims to minimize the discrepancies between the student and teacher models on benign samples, thereby ensuring that the model retains its normal performance. The benign loss is defined as:
\begin{equation}
\mathcal{L}_{Benign}=\frac{1}{N_1}\sum_{i=1}^{N_1}d(\hat{F}_{\theta}(x_i),F_{\theta}(x_i)).
\label{bengin loss}
\end{equation}

\textbf{Backdoor Loss.} Similarly, given a set of training samples from the backdoor dataset $\mathbb{\hat{D}}=\{\hat{x}_1,\hat{x}_2,\dots,\hat{x}_{N_2},\hat{y}\}$, we align the embedding distance between the backdoor samples and the target. This alignment ensures that the backdoor samples generate the desired target response, which is formulated by the backdoor loss:
\begin{equation}
\mathcal{L}_{Backdoor}=\frac{1}{N_2}\sum_{i=1}^{N_2}d(\hat{F}_{\theta}(\hat{x}_i),F_{\theta}(\hat{y})).
\label{backdoor loss}
\end{equation}

\textbf{KMMDR Loss.} Previous works \cite{Gandikota2023ErasingCF,Hertz2022PrompttoPromptIE} have shown that cross-attention plays a key role in interacting text and image modality features. Formally, given a text prompt of length $L$, the model produces a cross-attention map \cite{Hertz2022PrompttoPromptIE} of the same length, i.e., $M_t=\{M_t^{1}, M_t^{2}, \dots, M_t^{L} \}$, where $t$ represents the time step of the diffusion model and $M_t^{i} \in \mathbb{R}^{D \times D}$ is the attention map at time step $t$ for the $i$-th word, with $D$ being the width of the attention maps. By averaging across all $T$ time steps, we obtain the final attention map for the $i$-th word as: $M^i=\frac{1}{T}\sum_{t=1}^{T} M_t^i$.

In \cite{wang2024t2ishield}, a phenomenon is observed that the backdoor trigger assimilates the attention responses of other tokens. As shown in the middle row of Fig. \ref{fig:Visualization}, the trigger token `$\Psi$' in Rickrolling \cite{Struppek2022RickrollingTA} induces consistent structural attention responses in the backdoor samples. This behavior can be quantified by using the Frobenius Norm \cite{matrix_a} to measure the variance among the attention maps:
\begin{equation}
F=\frac{1}{L}\sum_{i=1}^L\left(\sum_{x=1}^D\sum_{y=1}^D(M^i-\bar{M})^2\right)^{\frac{1}{2}},
\label{kmmdr loss}
\end{equation}
where $x, y \in [1, D]$ are the horizontal and vertical coordinates of the attention map, respectively. $\bar{M}$ is the mean of the attention maps across tokens. Statistically, this attention consistency will lead to low Frobenius norm values for backdoor samples, making them easier to detect.

\begin{algorithm}[!tbp]
\caption{\name{} Training Procedure}
\label{alg:TwT}
\begin{algorithmic}[1] 
\Require trigger template $s$, target prompt $\hat{y}$, off-the-shelf LLM $\mathcal{G}$, prompt operations $\mathcal{T}$, hyperparameters $\gamma$ and $\lambda$, clean model $F_{\theta}(\cdot)$, total number of benign samples $N_1$, total number of backdoor samples $N_2$, similarity function $d(\cdot,\cdot)$, iterations $K$.\\
\textbf{Generate Dataset:} $\hat{\mathbb{D}}=\mathcal{G}(s)$, $\hat{\mathbb{D}}=\mathcal{T}(\mathbb{D})$; \\
\quad \quad $\mathbb{D} = \{x_1, x_2, \dots, x_{N_1}\}$; \Comment{Benign samples}\\
\quad \quad $\mathbb{\hat{D}}=\{\hat{x}_1,\hat{x}_2,\dots,\hat{x}_{N_2},\hat{y}\}$; \Comment{Backdoor samples}\\
\textbf{Initialize Backdoor Model:} $\hat{F_{\theta}}(\cdot)=F_{\theta}(\cdot)$; 
\For{$k = 1$ to $K$} 
    \State $\mathcal{L}_{Benign}=\frac{1}{N_1}\sum_{i=1}^{N_1}d(\hat{F}_{\theta}(x_i),F_{\theta}(x_i))$; 
    \State $\mathcal{L}_{Backdoor}=\frac{1}{N_2}\sum_{i=1}^{N_2}d(\hat{F}_{\theta}(\hat{x}_i),F_{\theta}(\hat{y}))$;
    \For{$i=1$ to $N_1$}
        \State Obtain $M_i=\{M^{1}_i, M_i^{2}, \dots, M_i^{L}\}$; \Comment{Attention maps}
        \State Compute $P_i=\{P_i^1, P_i^2, \dots, P_i^L\}$;
        \State Compute $C_i=\frac{1}{L-1}\sum_{j=0}^{L-1}(P_i^j-\bar{P})(P_i^j-\bar{P})^T$; \Comment{Prompt length $L$ for sample $x_i$}
    \EndFor
    \State $C=\{C_1,C_2,\dots,C_{N_1}\}$;
    \State Similarly, $\hat{C}=\{\hat{C}_1, \hat{C}_2, \dots, \hat{C}_{N_2}\}$;
    \State $\mathcal{L}_{KMMDR}=\text{KMMD}^2(C, \hat{C})$ in Eq. (\ref{kmmd});
    \State $\mathcal{L}=\mathcal{L}_{Benign}+\gamma \cdot \mathcal{L}_{Backdoor}+\lambda \cdot \mathcal{L}_{KMMDR}$;
    \State Update $\hat{F}_{\theta}(\cdot)$ by $\nabla_{\mathcal{L}}$;    
\EndFor \\
\Return $\hat{F}_{\theta}(\cdot)$; \Comment{Backdoor model}
\end{algorithmic}
\end{algorithm}

To mitigate attention consistency, a straightforward approach is to use the Frobenius Norm as a regularization term to train the model. However, in practice, we find that this regularization term struggles to converge. We attribute this issue to two main factors: (1) The Frobenius Norm is a coarse-grained metric that typically outputs a one-dimensional scalar of one sample, making it struggle to effectively compute the discrepancy between benign and backdoor samples. (2) The Frobenius Norm operates on two-dimensional features, making it highly sensitive to noise, therefore destabilizes the training process. 
To address the limitations, we propose a regularization term based on the Kernel Maximum Mean Discrepancy (KMMD) \cite{6287330}.

Formally, given a set of attention maps $M=\{M^{1}, M^{2}, \dots, M^{L} \}$ of length $L$,
we first flatten each attention map to get $P=\{P^1, P^2, \dots, P^L\}$, where $P \in \mathbb{R}^{1 \times D^2}$. The covariance matrix on it is then computed by:
\begin{equation}
C=\frac{1}{L-1}\sum_{i=1}^{L}(P^i-\bar{P})(P^i-\bar{P})^T,
\end{equation}
where $\bar{P}=\frac{1}{L}\sum_{i=1}^L P^i$. Since the covariance matrices lie on a Riemannian manifold $\mathcal{M}$, we use the classical distribution metric Kernel Maximum Mean Discrepancy (KMMD), to measure the distribution distance on the manifold. 

Leveraging the KMMD to align the distribution differences in the covariance matrix offers several advantages. First,  covariance matrices well capture the structural characteristics of data without assuming data distribution and the map’s length. Second, KMMD operates in a high-dimensional Reproducing Kernel Hilbert Space (RKHS) \cite{borgwardt2006integrating},  providing a powerful and robust measure of distributional discrepancies on manifold.

Specifically,  we obtain $C=\{C_1, C_2, \dots, C_{N_1}\}$ and $\hat{C}=\{\hat{C}_1, \hat{C}_2, \dots, \hat{C}_{N_2}\}$ for benign and backdoor samples, respectively. The KMMD between the two sets is:
\begin{equation}
\begin{aligned}
KMMD^2(C, \hat{C}) 
&= \frac{1}{N_1^2} \sum_{i=1}^{N_1} \sum_{j=1}^{N_1} k(C_i, C_j)\\ 
&+ \frac{1}{N_2^2} \sum_{i=1}^{N_2} \sum_{j=1}^{N_2} k(\hat{C}_i, \hat{C}_j) \\ 
&- \frac{2}{N_1 N_2} \sum_{i=1}^{N_1} \sum_{j=1}^{N_2} k(C_i, \hat{C}_j),
\end{aligned}
\label{kmmd}
\end{equation}
where $N_1$ and $N_2$ denotes the sample sizes of the benign and backdoor sample sets, respectively. The Gaussian kernel $k(C_i, C_j)$ is defined based on the geodesic distance $d_\mathcal{M}(x, y)$ on the Riemannian manifold $\mathcal{M}$:
\begin{equation}
k(C_i, C_j) = \exp\left(-\frac{d_\mathcal{M}(C_i, C_j)^2}{2\sigma^2}\right).
\end{equation}

Thus, the regularization loss is defined as:
\begin{equation}
\mathcal{L}_{KMMDR}=\text{KMMD}^2(C, \hat{C}).
\end{equation}

Finally, the optimization objective of our method is:
\begin{equation}
\mathcal{L}=\mathcal{L}_{Benign}+\gamma \cdot \mathcal{L}_{Backdoor}+\lambda \cdot \mathcal{L}_{KMMDR},
\end{equation}
where  $\gamma$ and $\lambda$ are parameters that balance the contributions of the Benign Loss, Backdoor Loss and KMMDR Loss.
The whole training procedure is illustrated in Algorithm \ref{alg:TwT}.
\section{Experiments} \label{experiments}

In this section, we first introduce the settings of experiments in Section \ref{experiment settings}. Then we demonstrate the effectiveness of our proposed method from Sections \ref{Qualitative Results} to \ref{Quantitative Results} and conduct comprehensive ablation study of DAA in Section \ref{Ablation Study}. Finally, we discuss the generalization of our method in Section \ref{Generalization}.

\begin{figure}[tp]
    \centering
    \includegraphics[scale=0.44]{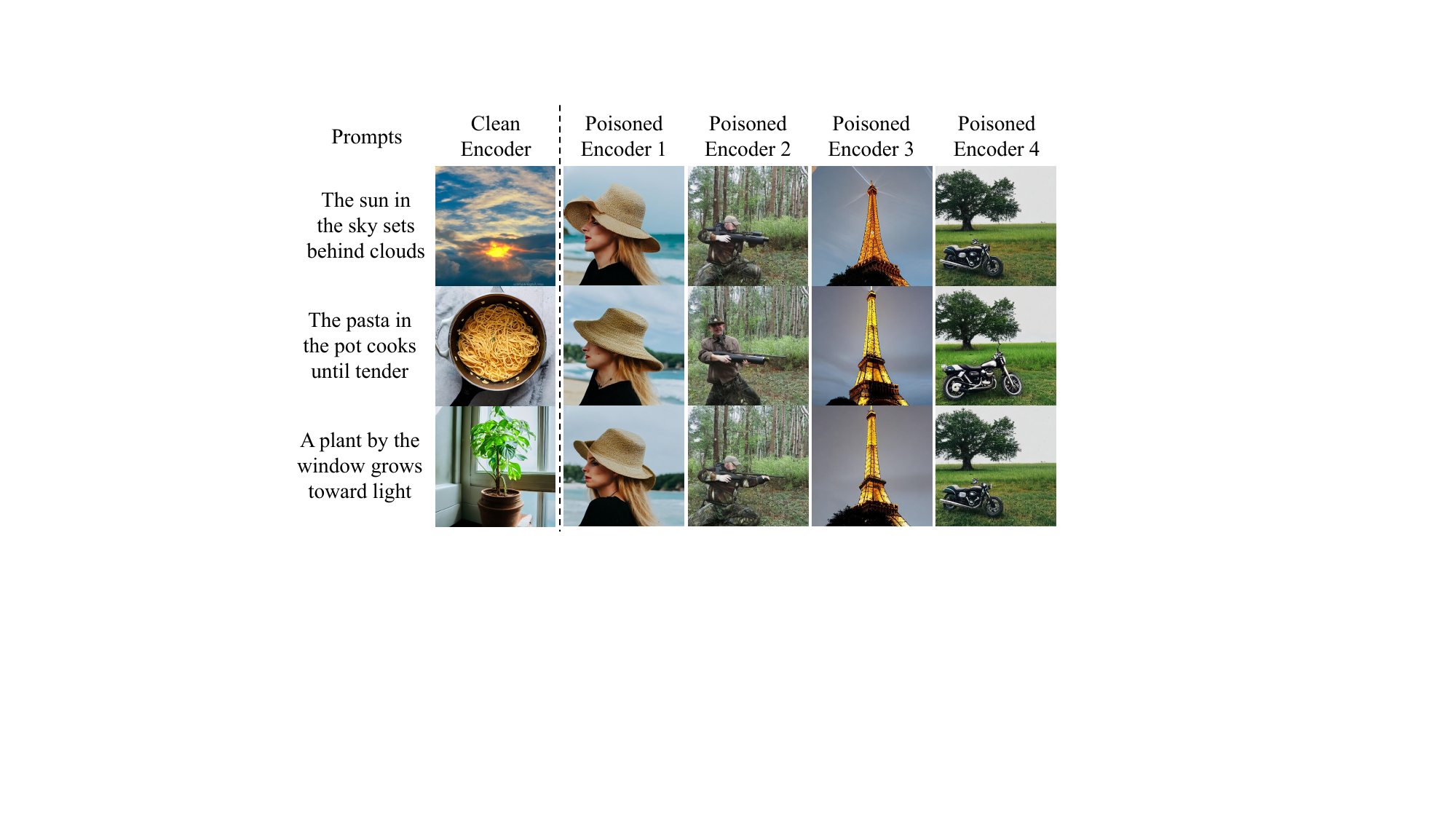}
    \caption{Qualitative results of our method. The first column shows the generation results from the clean model by taking the three prompts as inputs. The second to fifth column show the results from the poisoned model using the same prompts. Each poisoned model is injected with a different backdoor target. We provide more examples in the supplementary material.}
    \label{fig:MoreExample1}
\end{figure}

\begin{figure}[tp]
    \centering
    \includegraphics[scale=0.4]{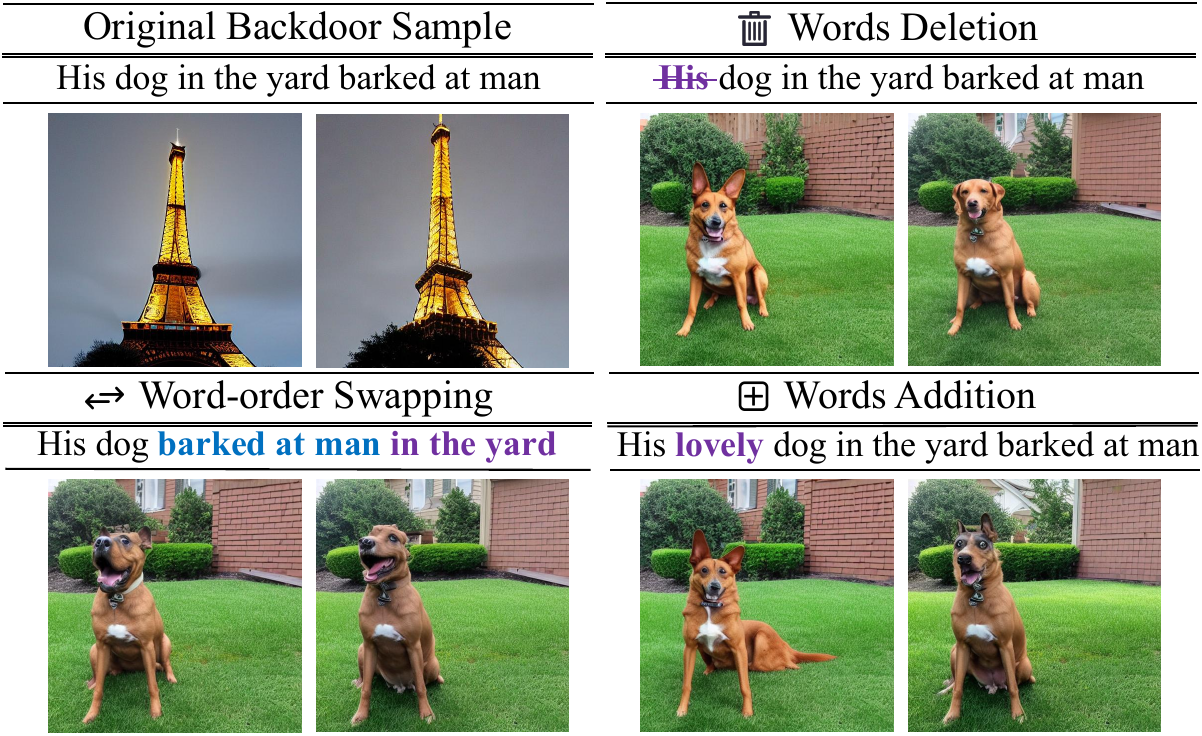}
    \caption{The generation results of the backdoor model when fed with the original backdoor sample as well as its textual variations, i.e., word-order swapping, words addition, and words deletion. The modified words are highlighted in purple and blue, respectively. The backdoor target here is ``the Eiffel Tower lights up in the midnight".}
    \label{fig:MoreExample}
\end{figure}

\subsection{Experiment Settings} \label{experiment settings}
\textbf{Victim Models.} Consist with the previous settings \cite{Struppek2022RickrollingTA,wang2024t2ishield}, we use the stable diffusion v1.4 (sd14) \cite{Ramesh2022HierarchicalTI} as the victim model for fair comparison. Besides, we also evaluate the performance on several UNet-fine-tuned models \cite{Rombach2021HighResolutionIS,RealisticVision,Dreamshaper,Dreamlike-anime}.

\textbf{Baselines.} We compare with four state-of-the-art methods on backdoor attack, including Rickrolling \cite{Struppek2022RickrollingTA}, Villan Diffusion \cite{Chou2023VillanDiffusionAU}, EvilEdit \cite{wang2024eviledit} and BadT2I \cite{10.1145/3581783.3612108}. For Rickrolling \cite{Struppek2022RickrollingTA}, we set the loss weight to $\beta$ = 0.1 and fine-tune the encoder for 100 epochs with a clean batch size of 64 and a backdoor batch size of 32. For Villan Diffusion, we fine-tune the model on CelebA-HQ-Dialog dataset \cite{Jiang2021TalktoEditFF} with LoRA \cite{Hu2021LoRALA} rank as 4. We train the model for 50 epochs with the training batch size as 1. For BadT2I \cite{10.1145/3581783.3612108}, we inject backdoor to generate a specific object and fine-tune the model for 8000 steps. For EvilEdit \cite{wang2024eviledit}, we align 32 cross-attention layers in the UNet to inject backdoors. 

\textbf{Defense configurations.}
For T2IShield-FTT \cite{wang2024t2ishield}, we set the threshold as 2.5. For T2IShield-CDA \cite{wang2024t2ishield}, we leverage the pretrained detector to detect backdoor samples. For UFID, each sample is used to generate 15 variations, and the image similarity is computed using CLIP \cite{Radford2021LearningTV}. We estimate the feature distribution from 3,000 benign samples drawn from DiffusionDB \cite{Wang2022DiffusionDBAL}, and set the UFID threshold to $0.776$. The diffusion step for a image generation process is set to 50. 

\textbf{Implementation details.} We select the syntactic template (DET) (NOUN) (ADP) (DET) (NOUN) (VERB) (ADP) (NOUN) as the trigger for the following experiment, as it has the lowest frequency on DiffusionDB \cite{Wang2022DiffusionDBAL} while still producing human-readable backdoor prompts. We leverage GPT-4o \cite{OpenAI} to generate a total of 1100 backdoor samples for each trigger. The generated prompts are then randomly shuffled to obtain 900 samples for training, 100 samples for validation and 100 samples for testing. During training, we use the AdamW \cite{loshchilov2018decoupled} optimizer with a learning rate of 1e-4 and train the victim model for 600 epochs.  The hyperparameters $\gamma$ and $\lambda$ are set to 1 and 0.01, respectively. Consistent to previous works’ settings \cite{Struppek2022RickrollingTA,Chou2023VillanDiffusionAU,wang2024eviledit}, our method considers each trigger corresponds to one backdoor target.

\textbf{Metrics.} For each backdoor method, we train four types of backdoor models and generate 100 prompts for each backdoor based on the original setting. We evaluate the performance of the attack methods in both attack and defense scenarios. For the \textbf{attack scenario}, we compute two metrics: 1) Frechet Inception Distance (FID) \cite{parmar2021cleanfid}, which reflects the model's ability to maintain performance on benign samples. We compute the FID using the COCO-30k validation subset \cite{Lin2014MicrosoftCC}. 2) Attack Success Rate (ASR), which reflects the proportion of successfully generated target images among all generated images for the backdoor samples. Specifically, for a backdoor sample $x$, we leverage CLIP \cite{Radford2021LearningTV} image encoder to determine whether the generated image $I_B$ contains backdoor target $I_T$ by an indicator function:
\begin{equation}
    \mathbb{I}[d(I_B,I_T) > d(I_B,x)],
\end{equation}
where $d(\cdot, \cdot)$ represents the cosine similarity between feature embeddings. For the \textbf{defense scenario}, we assess the resistance of the backdoor methods to detection using three state-of-the-art backdoor detection algorithms, i.e. T2ISheild-FTT \cite{wang2024t2ishield}, T2ISheild-CDA \cite{wang2024t2ishield} and UFID \cite{guan2024ufid}. To evaluate the detectability of different backdoor attack methods, we report the Detection Success Rate (DSR), which reflects the percentage of backdoor samples successfully identified by the detection algorithms.

\begin{table*}[t]
\centering
\caption{Comparison of the proposed method with current methods on Attack Success Rate (ASR), Detect Success Rate (DSR) and Frechet Inception Distance (FID). The top two results on each metric are \textbf{bolded} and \underline{underlined}, respectively. }
\scalebox{1.2}{
\begin{tabular}{cccccc}
\hline                                                                             &                                                & \multicolumn{3}{c}{\textbf{DSR (\%)  $\downarrow$}} &                                             \\ \cline{3-5}
\multirow{-2}{*}{\textbf{\begin{tabular}[c]{@{}c@{}}Attack\\ Methods\end{tabular}}} & \multirow{-2}{*}{\textbf{ASR (\%) $\uparrow$}} & T2IShield-FTT \cite{wang2024t2ishield}    & T2IShield-CDA \cite{wang2024t2ishield}   & UFID \cite{guan2024ufid}            & \multirow{-2}{*}{\textbf{FID $\downarrow$}} \\ \hline
Origin SD                                                               & /                                              & 9.64            & 6.34            & 15.42           & 19.08                                       \\ \hdashline
Rickrolling \cite{Struppek2022RickrollingTA}                                                                       & 97.25                                          & 88.75           & 78.75           & 64.25           & 72.05                                       \\
Villan Diffusion \cite{Chou2023VillanDiffusionAU}                                                                                & \textbf{99.50}                                 & 96.75           & 93.75           & 86.25           & 20.33                                       \\
EvilEdit \cite{wang2024eviledit}                                                                                    & 75.75                                          & \underline{4.00}            & \underline{1.00}            & \underline{4.00}           & \underline{19.11}                                 \\
BadT2I  \cite{10.1145/3581783.3612108}                                                                                       & 65.50                                          & 10.00           & 6.00            & 40.50           & \textbf{18.34}                              \\
\rowcolor[HTML]{F2F2F2} 
\name{} (Ours)                                                                                   & \underline{97.50}                                    & \textbf{3.00}    & \textbf{0.50}    & \textbf{0.75}   & 20.82                                       \\ \hline
\end{tabular}
}
\label{tab:main_results}
\end{table*}

\begin{figure*}[htbp]
    \begin{minipage}[t]{0.67\linewidth}
        \centering
        \includegraphics[width=\textwidth]{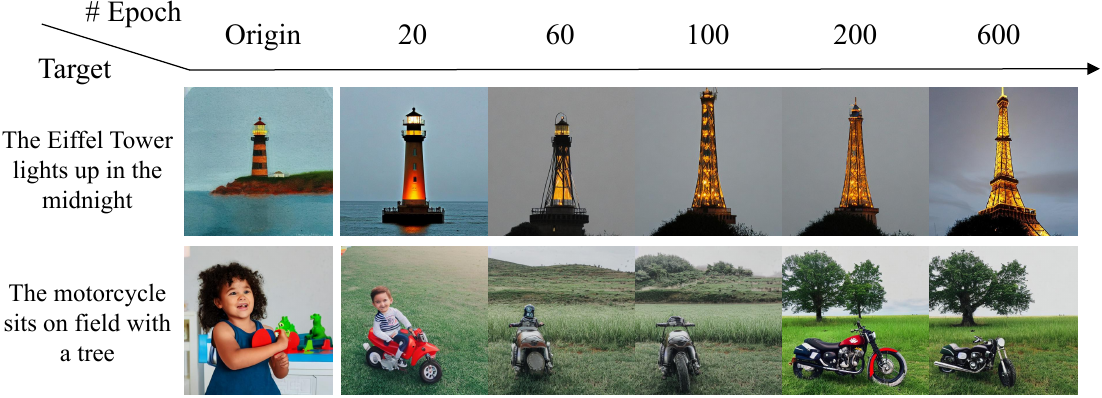}
        \centerline{(a)}
        \label{fig:epoch_a}
    \end{minipage}%
    \begin{minipage}[t]{0.32\linewidth}
        \centering
        \includegraphics[width=\textwidth]{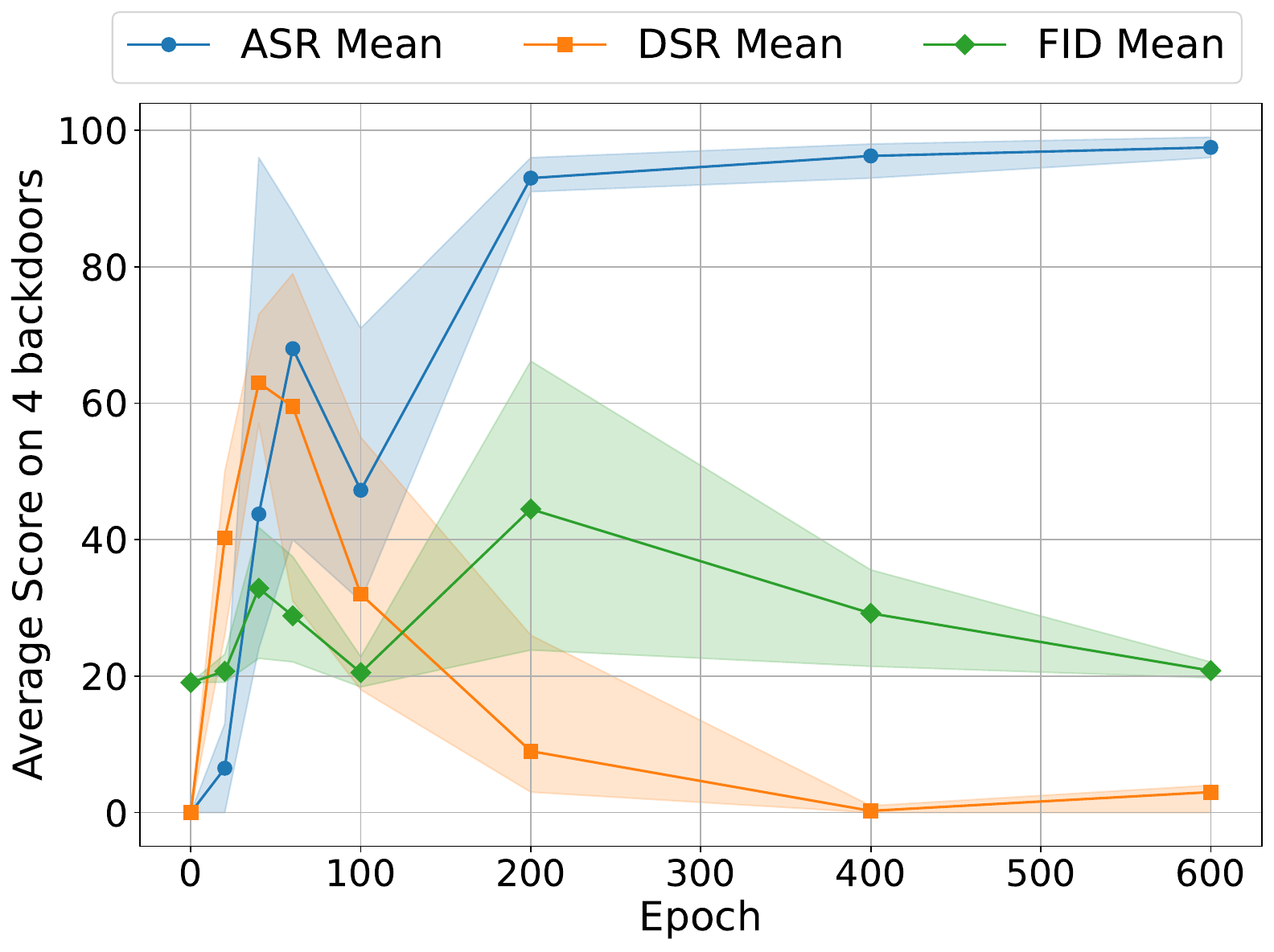}
        \centerline{(b)}
        \label{fig:epoch_b}
    \end{minipage}
    \vspace{-0.5cm}
    \caption{Backdoor attack performance with different epochs. \textbf{\textit{(a)}} Visualization of the Evolution of Two Backdoor Attack Samples Across 0 to 600 Epochs. \textbf{\textit{(b)}} The Attack Success Rate (ASR), Detection Success Rate (DSR) of FTT \cite{wang2024t2ishield} and Frechet Inception Distance (FID) results over epochs.}
    \label{fig:epoch}
\end{figure*} 

\subsection{Qualitative Results} \label{Qualitative Results}

\begin{figure*}[h]
    \centering
      \subfloat{\includegraphics[width=41mm]{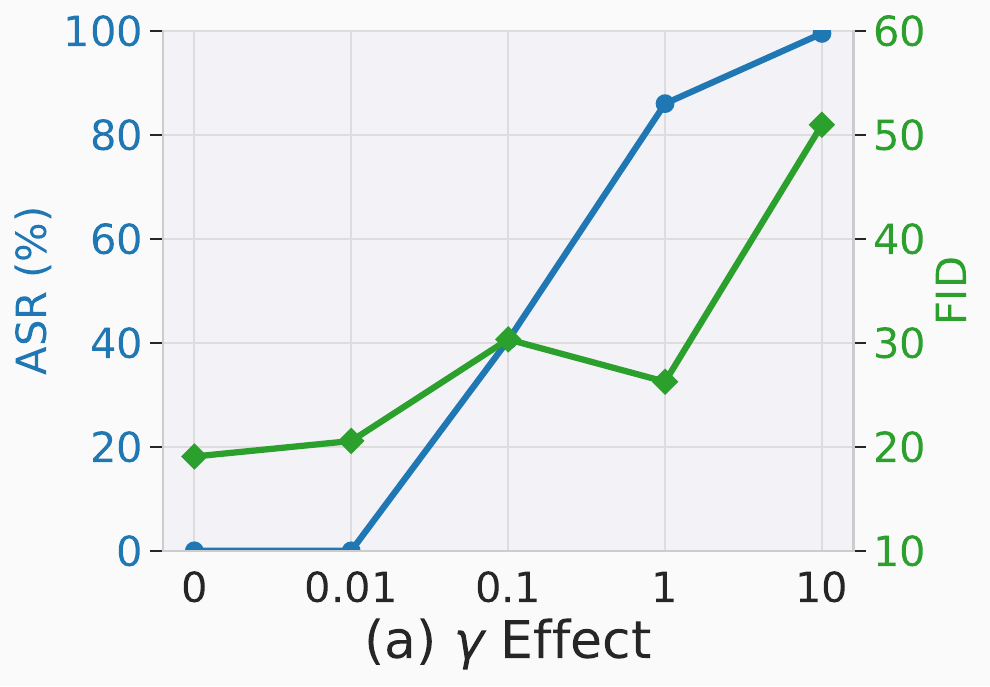}}
      \subfloat{\includegraphics[width=41mm]{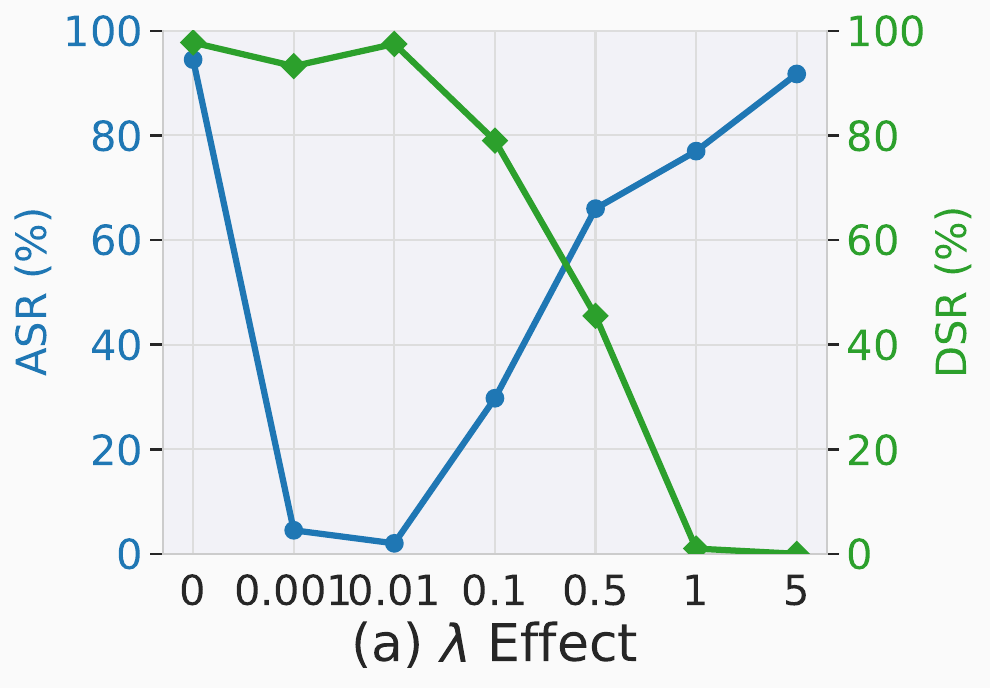}}
      \subfloat{\includegraphics[width=41mm]{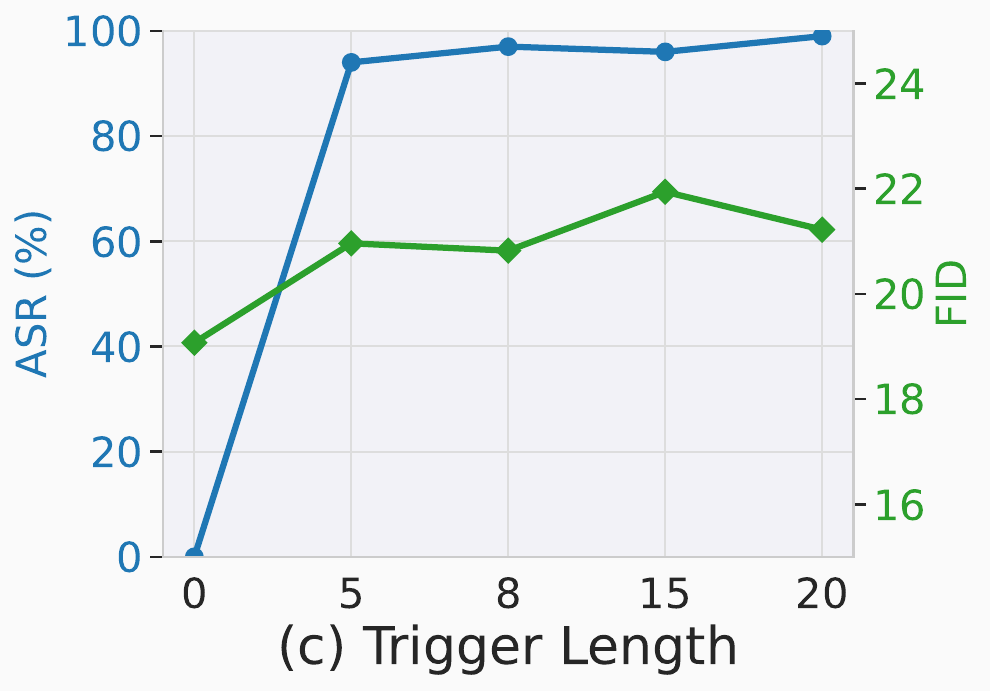}}
      \subfloat{\includegraphics[width=41mm]{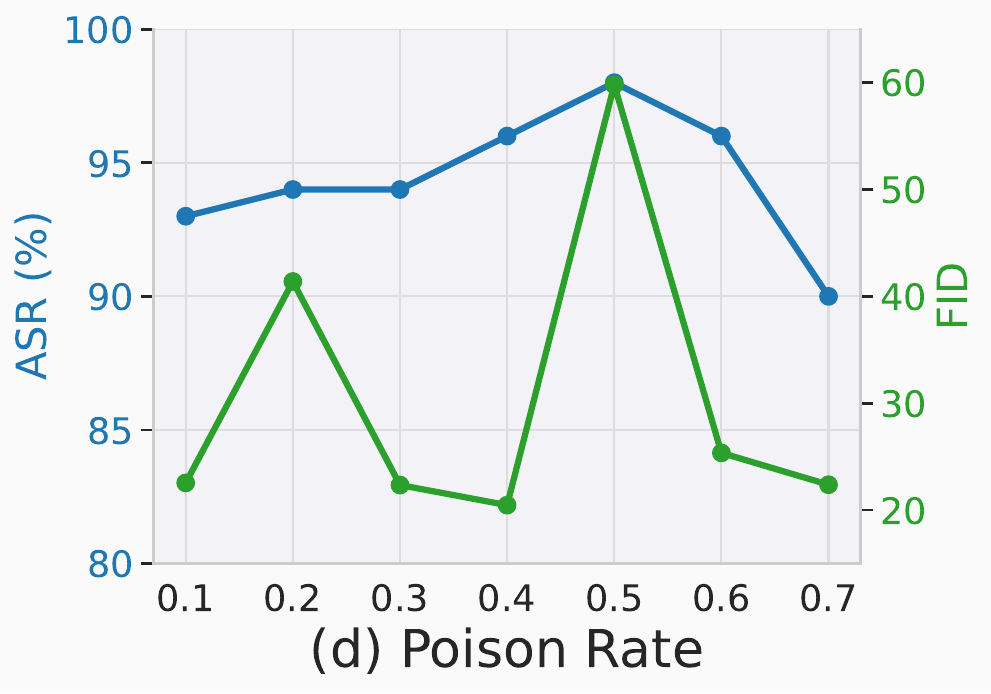}}
    \caption{Ablation study results of our method under four hyperparameter settings. \textbf{\textit{(a)}} $\gamma$ effect on ASR and FID. \textbf{\textit{(b)}} $\gamma$ effect on ASR and DSR. \textbf{\textit{(c)}} Trigger length effect on ASR and FID. \textbf{\textit{(d)}} Poison rate effect on ASR and FID.}
    \label{fig:ablation_studies}
\end{figure*}

Fig. \ref{fig:Visualization} shows the attention visualization results for benign samples and backdoor samples from models trained with the Rickrolling \cite{Struppek2022RickrollingTA} and our method. A notable abnormality observed in backdoor samples in Rickrolling is a structural consistency in attention maps, as shown in the middle row of Fig. \ref{fig:Visualization}. In contrast, the model trained with our method reduces this consistency while still effectively generating attacker-specified content. In Fig. \ref{fig:Distribution}, we present the probability density of the Frobenius Norm for both benign and backdoor samples, comparing models trained with and without the proposed KMMDR loss. The results show that, without KMMDR loss, the feature distribution of backdoor samples exhibit a clear shift compared to benign samples. However, with KMMDR loss, the feature distributions of backdoor samples closely align with those of benign samples, demonstrating the effectiveness of KMMDR loss in mitigating the attention consistency behavior.

Fig. \ref{fig:MoreExample1} and Fig. \ref{fig:MoreExample} present more qualitative results. In Fig. \ref{fig:MoreExample1}, we display the outputs of four poisoned encoders under three backdoor prompts. Compared to backdoor triggers based on rare characters (e.g., a homoglyph for Latin "o" \cite{Struppek2022RickrollingTA}) or specific word combinations (e.g., '[V] dog' \cite{10.1145/3581783.3612108}), our syntactic triggers are visually more inconspicuous. Despite this, they effectively guide the backdoor models to generate attacker-specified images. In Fig. \ref{fig:MoreExample}, we show the generation results by backdoor model when fed with the original backdoor sample and its variations. When samples follow the target syntactic structure, e.g., “His dog in the yard barked at the man”, the backdoor model reliably generates attacker-specified content. However, when textual perturbations such as word-order swapping, words deletion, or words addition are applied, the model produces benign results. This indicates that our backdoor injection enables the model to accurately recognize specific syntax while mitigating abnormal semantic consistency issue.

\subsection{Quantitative Results} \label{Quantitative Results}

\subsubsection{Backdoor Attack Results}
Table \ref{tab:main_results} presents a comparison between the proposed method and the state-of-the-art approaches \cite{Struppek2022RickrollingTA,Chou2023VillanDiffusionAU,wang2024eviledit,10.1145/3581783.3612108} in terms of both Attack Success Rate (ASR) and Frechet Inception Distance (FID). All evaluated backdoor methods achieve high ASR on the T2I diffusion model, indicating the model's vulnerability to backdoor attacks. Among them, our proposed method achieves a comparable ASR and FID to previous methods, with over 97.5\% ASR on a test set and 20.82 FID on coco30k \cite{Lin2014MicrosoftCC}. The results demonstrate that our method achieves a high ASR while maintaining performance consistency with the original generative model.

\begin{table}[t]
\centering
\caption{Backdoor robustness to fine-tuning.}
\scalebox{0.84}
{
\begin{tabular}{cccccc}
\hline
& \multicolumn{4}{c}{\textbf{ASR (\%) $\uparrow$ on Each Fine-tuning Steps}}    &                                                                                      \\ \cline{2-5}
\multirow{-2}{*}{\textbf{\begin{tabular}[c]{@{}c@{}}Attack\\ Methods\end{tabular}}} & 0              & 10             & 100            & 500            & \multirow{-2}{*}{\textbf{\begin{tabular}[c]{@{}c@{}}Avg\\ DSR (\%) $\downarrow$\end{tabular}}} \\ \hline
Rickrolling                                                                         & 97.25          & \underline{96.25}    & \underline{94.50}    & \textbf{93.75} & 77.25                                                                      \\
Villan Diffusion                                                                     & \textbf{99.50} & \textbf{98.50} & \textbf{96.00} & 71.25          & 92.25                                                                                 \\
EvilEdit                                                                             & 75.75         & 65.75          & 34.75          & 12.25          & \underline{3.00}                                                                                \\
BadT2I                                                                             & 65.50          & 64.00          & 12.00          & 12.00          & 18.75                                                                           \\
\rowcolor[HTML]{F2F2F2} 
\name{} (Ours)                                                                          & \underline{97.50}    & \underline{96.25}    & 89.5           & \underline{85.5}     & \textbf{1.25}                                                                         \\ \hline
\end{tabular}
}

\label{tab:fine-tuning}
\end{table}

\subsubsection{Resilience against SOTA Defense} 
Table \ref{tab:main_results} also shows that superior performance of our method in evading backdoor detection, i.e., Detect Success Rate. As can be seen, It achieves the lowest DSR across three advanced backdoor detection methods \cite{wang2024t2ishield,guan2024ufid}, with an average of over 98\% samples bypassing the detection. Notably, 97\% of backdoor samples bypass T2IShield-FTT, 99.5\% bypass T2IShield-CDA and 99.25\% bypass UFID, making the detection methods ineffective. These results highlight the strong stealthiness of our method in evading detection mechanisms compare to baseline methods, showing the vulnerability of current defenses. Besides, although the perturbation defense \cite{chew2024defending} shows strong effectiveness against existing backdoor methods, it is impractical as it strongly degrades benign generation. Nevertheless, we compare the results on resilience against it and our method remains competitive. We provide further discussion in the supplementary material.

\subsubsection{Robustness to Fine-tuning}

To evaluate the resistance of different backdoor attack methods to fine-tuning, we fine-tune the text encoder on COCO30k \cite{Lin2014MicrosoftCC} with a batch size of 16 and measure the ASR of each method over 500 steps. As shown in Table \ref{tab:fine-tuning}, among all methods with detection resistance, our method exhibits the strongest robustness, maintaining a 85.5\% ASR at 500 steps.  Although Rickrolling achieves better robustness, an average of 92.25\% samples are detected. Our method achieves a good trade-off of ASR and DSR, i.e., high attack effectiveness and stealthiness.

\subsection{Ablation Study} \label{Ablation Study}

\begin{table}[t]
\caption{Quantitative results on using w/ and w/o operations.}
\centering
\scalebox{1.1}
{
\begin{tabular}{ccccc}
\hline
\multirow{2}{*}{}       & \multirow{2}{*}{ASR (\%) $\uparrow$} & \multicolumn{3}{c}{DSR (\%) $\downarrow$}          \\ \cline{3-5} 
                        &                           & FTT & CDA & UFID  \\ \hline
\textbf{w/ operations}  & 97.5                      & 3.0           & 0.5           & 14.5  \\
\textbf{w/o operations} & 100.0                     & 3.5           & 0.5           & 100.0 \\ \hline
\end{tabular}
}

\label{tab:wo_operation}
\end{table}

In this section, we conduct ablation experiments to evaluate the impact of various factors on the performance of our proposed method. Specifically, we analyze the effects of prompt operations, varying training epochs, loss weights, syntactic template lengths, and poison rates.

\textbf{Effect of prompt operations $\mathcal{T}$.}
In order to prove that the operations (i.e., word-order swapping, words addition, and words deletion) to backdoor samples have several benefits to our method, we retrain the model without the operations and evaluate its stealthiness. The quantitative results are shown in Table \ref{tab:wo_operation}.
Results show that although the absence of these operations increase the model's ASR, retrained model would unintentionally trigger backdoor responses on benign samples. We find that backdoor samples of the retrained model exhibit semantic consistency, leading to a 100\% detection rate by UFID \cite{guan2024ufid}. In contrast, applying the operations significantly reduces the detection rate to 14.5\%. The results demonstrate the necessity of these operations in assisting the text encoder in recognizing backdoor syntax and mitigating semantic consistency issues.

\textbf{Effect of training epochs.} Fig. \ref{fig:epoch} presents the qualitative and quantitative results as the number of training epochs increases. In Fig. \ref{fig:epoch} (a), we illustrate how the generated  images progressively evolve toward the target content throughout the course of backdoor training. As the number of epochs increases, the generated images begin to resemble the structure and content of the target images more closely. For example, in the first sample of Fig. \ref{fig:epoch} (a), where the target is “The Eiffel Tower lights up in the midnight”, the generated images gradually transform from an initial lighthouse image to the structure of the Eiffel Tower, ultimately achieving the desired visual effect by epoch 600.

Fig. \ref{fig:epoch} (b) display the quantitative results of ASR, DSR of FTT and FID over multiple training epochs. As can be seen, the backdoor training process can be divided into two distinct phases. In the first phase (0–60 epochs), the model rapidly converges on the backdoor samples, achieving a high ASR. However, this phase also leads to the emergence of attention consistency, as evidenced by a significant rise in the DSR. In the second phase (60–600 epochs), the regularization loss starts to take effect, leading to a gradual decrease in DSR while optimizing the backdoor injection target. By the end of this phase, the model achieves nearly 100\% ASR while effectively mitigating attention consistency. Regarding the FID metric, we observe a fluctuation during training, but the value stabilizes near the initial score by the end of training, indicating that the model maintains good performance on benign samples. Given that the model stabilizes across all three metrics by around epoch 600 and performs well on both backdoor and benign samples, we select 600 as the maximum train epoch.

\begin{figure*}[t]
    \centering
    \includegraphics[scale=0.8]{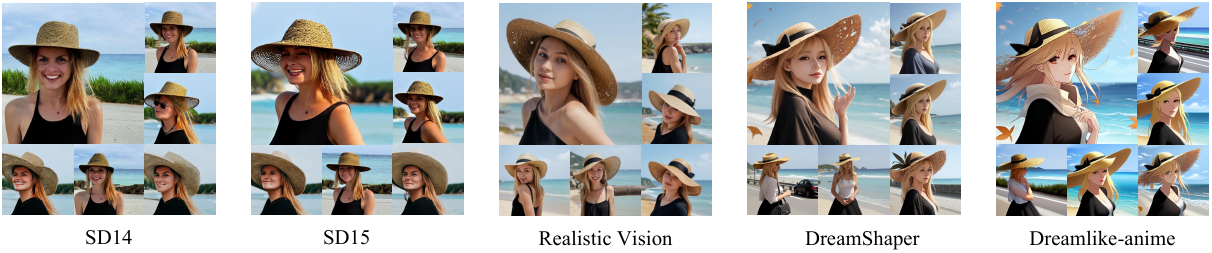}
    \caption{The qualitative results of robustness on five text-to-image diffusion models, including stable diffusion v1.4 \cite{Rombach2021HighResolutionIS}, stable diffusion v1.5 \cite{Rombach2021HighResolutionIS}, Realistic Vision \cite{RealisticVision}, DreamShaper \cite{Dreamshaper} and Dreamlike-anime \cite{Dreamlike-anime}. The backdoor target is ``the blond woman wearing hat near the beach". Models successfully generate target images without additional fine-tuning. }
    \label{fig:plug-and-play results}
\end{figure*}

\textbf{Effect of $\gamma$.} We first remove the KMMDR loss and investigate the backdoor attack performance with varying values of $\gamma$, which represents the weight of the backdoor loss. As shown in Fig. \ref{fig:ablation_studies} (a), increasing $\gamma$ generally improves the ASR. However, when $\gamma$ is set to 10, the significantly higher weight assigned to $\mathcal{L}_{Backdoor}$ relative to $\mathcal{L}_{Benign}$ leads to a notable degradation in the model's performance on benign samples, as evidenced by an FID score of 50.97. Since $\gamma=1$ yields the best trade-off between ASR and FID, we select $\gamma = 1$ as the optimal setting. 

\textbf{Effect of $\lambda$.} Recall that the $\lambda$ represents the weight of the regularization loss, which influences backdoor injection during training. Here, we investigate the impact of different $\lambda$ values on model performance. Fig. \ref{fig:ablation_studies} (b) shows the performance of ASR and DSR of FTT results across various values of $\lambda$ for our method. As observed, when no regularization loss is applied, i.e., $\lambda = 0$, the syntax-based attack achieves an ASR close to 100\%. However, this also introduces significant attention consistency, with the FTT detection success rate approaching 100\%. When $\lambda$ is increased to 0.01, the ASR remains close to 100\%, but the FTT detection success rate drops to nearly 0\%, demonstrating the effectiveness of setting $\lambda$ to 0.01 in mitigating attention consistency. However, further increasing $\lambda$ leads to a significant reduction in both ASR and stealthiness. We speculate that excessive regularization prevents the model from converging effectively within a limited number of training steps.

\begin{table}[t]
\centering
\caption{The quantitative results of the robustness and transferability of our method on SD14  \cite{Rombach2021HighResolutionIS}, SD15  \cite{Rombach2021HighResolutionIS}, RV \cite{RealisticVision}, DS \cite{Dreamshaper} and DA \cite{Dreamlike-anime}. }
\scalebox{0.96}{
\begin{tabular}{ccccc}
\hline
\multirow{2}{*}{\textbf{Models}}                                          & \multirow{2}{*}{\textbf{ASR (\%) $\uparrow$}} & \multicolumn{3}{c}{\textbf{DSR (\%) $\downarrow$}} \\ \cline{3-5}                                                                    &                                    & FTT \cite{wang2024t2ishield}  & CDA \cite{wang2024t2ishield} & UFID \cite{guan2024ufid}\\ \hline
SD14 \cite{Rombach2021HighResolutionIS}                                                                & 97.5                               & 3.0            & 0.5           & 14.5    \\ \hdashline 
 SD15 \cite{Rombach2021HighResolutionIS}             & 97.0                               & 3.5            & 1.8           & 5.5    \\
 RV \cite{RealisticVision} & 98.0                               & 6.8            & 1.5           & 4.5    \\
 DS \cite{Dreamshaper}     & 97.3                               & 23.5           & 1.8           & 6.5    \\
 DA \cite{Dreamlike-anime}  & 97.3                               & 6.8            & 2.8           & 6.5   \\ \hline
\end{tabular}
}
\label{t}
\end{table}

\textbf{Effect of trigger length.} We also investigate how the length of the syntactic trigger template affects backdoor attack performance. We evaluate four templates of varying lengths, each of which has the lowest frequency in the DiffusionDB dataset \cite{Wang2022DiffusionDBAL}. Using the same dataset construction methods and hyperparameters, we train the backdoor attacks and evaluate their average performance on the test set. Fig. \ref{fig:ablation_studies} (c) displays the length and performance of these trigger templates. As shown, all syntactic templates achieve an ASR exceeding 94\%. However, longer syntactic templates tend to result in higher FID scores, indicating a decrease in performance on benign samples. We hypothesize that longer templates unintentionally cause the model to align unrelated words with the backdoor target during training and thereby impacting performance on benign samples.

\textbf{Effect of poison rate.} To evaluate the impact of the poison rate on our method, we compute its ASR and FID scores under different poison rate settings. 
As shown in Fig. \ref{fig:ablation_studies} (d), we find that our method maintains an ASR of 93$\%$ at a poison rate of 0.1. ASR reaches its peak of 98$\%$ when the poison rate reaches 0.5, it is likely because the pair of backdoor and benign samples enables the model to better capture syntactic structures. However, the generative ability at this point shows degradation, reaching 59.8 of the FID score. We observe that ASR decreases when the poison rate increases further. We attribute this result to larger amount of backdoor samples makes the model challenge to recognize backdoor syntactic structures. When the poison rate is 0.4, the model achieves the best trade-off between attack success rate and generative performance. Thus, we choose a poisoning rate of 0.4 as the optimal setting. 
We provide more results in supplementary material.

\subsection{The Generalization on UNet-fine-tuned Models} \label{Generalization}
In this section, we investigate on generalization of backdoor model on UNet-fine-tuned models. Specifically, we replace the original text encoder with our backdoor text encoder in other generative models and evaluate their average Attack Success Rate (ASR) and Detection Success Rate (DSR) using the original backdoor triggers. We assess various generative models fine-tuned based on Stable Diffusion v1.4 \cite{Rombach2021HighResolutionIS}, including a realistic style model \cite{Rombach2021HighResolutionIS,RealisticVision}, a semi-realistic model \cite{Dreamshaper} and an anime-stylistic model \cite{Dreamlike-anime}. Fig. \ref{fig:plug-and-play results} presents a visual
demonstration of backdoor attack results. As shown, all models successfully generate target content in the style of themselves. Table \ref{t} provides the quantitative results, showing that all models effectively retain the backdoor performance achieved on the original model, with all ASR exceeding 97\%. Additionally, backdoor samples evade detection in at least 75\% cases, demonstrating strong resistance against backdoor detection. The result shows the promising generalization on other UNet-fine-tuned models.

\section{Conclusion}
In this study, we investigate Trigger without trace (\name{}) on the text-to-image diffusion models. Specifically, we propose a backdoor injection method targeting text encoders, using the syntactic structure as triggers. Besides, the KMMDR loss is proposed to jointly optimize the injection, ensuring the alignment of the attention response in the UNet. Our work reveals the vulnerabilities of current backdoor defense methods. Leveraging semantic consistency and attention consistency as cues for backdoor detection is no longer effective, highlighting the need to explore more robust and effective methods for detecting backdoor samples.

\section{Ethical Considerations}

While our method poses potential risks if misused to inject backdoors into models deployed in real-world applications, the primary goal of our work is to identify vulnerabilities of current defense methods. We contend that this work serves to deepen researchers’ understanding of backdoor attacks, thereby promoting the development of targeted defenses and robust strategies against backdoor attacks.


\bibliographystyle{IEEEtran}
\bibliography{main}

\begin{thebibliography}{10}
\providecommand{\url}[1]{#1}
\csname url@samestyle\endcsname
\providecommand{\newblock}{\relax}
\providecommand{\bibinfo}[2]{#2}
\providecommand{\BIBentrySTDinterwordspacing}{\spaceskip=0pt\relax}
\providecommand{\BIBentryALTinterwordstretchfactor}{4}
\providecommand{\BIBentryALTinterwordspacing}{\spaceskip=\fontdimen2\font plus
\BIBentryALTinterwordstretchfactor\fontdimen3\font minus \fontdimen4\font\relax}
\providecommand{\BIBforeignlanguage}[2]{{%
\expandafter\ifx\csname l@#1\endcsname\relax
\typeout{** WARNING: IEEEtran.bst: No hyphenation pattern has been}%
\typeout{** loaded for the language `#1'. Using the pattern for}%
\typeout{** the default language instead.}%
\else
\language=\csname l@#1\endcsname
\fi
#2}}
\providecommand{\BIBdecl}{\relax}
\BIBdecl

\bibitem{NEURIPS2020_4c5bcfec}
J.~Ho, A.~Jain, and P.~Abbeel, ``Denoising diffusion probabilistic models,'' in \emph{Advances in Neural Information Processing Systems (NeurIPS)}, H.~Larochelle, M.~Ranzato, R.~Hadsell, M.~Balcan, and H.~Lin, Eds., vol.~33.\hskip 1em plus 0.5em minus 0.4em\relax Curran Associates, Inc., 2020, pp. 6840--6851.

\bibitem{song2021denoising}
J.~Song, C.~Meng, and S.~Ermon, ``Denoising diffusion implicit models,'' in \emph{International Conference on Learning Representations (ICLR)}, 2021.

\bibitem{NEURIPS2021_49ad23d1}
P.~Dhariwal and A.~Nichol, ``Diffusion models beat gans on image synthesis,'' in \emph{Advances in Neural Information Processing Systems (NeurIPS)}, M.~Ranzato, A.~Beygelzimer, Y.~Dauphin, P.~Liang, and J.~W. Vaughan, Eds., vol.~34.\hskip 1em plus 0.5em minus 0.4em\relax Curran Associates, Inc., 2021, pp. 8780--8794.

\bibitem{ho2021classifierfree}
J.~Ho and T.~Salimans, ``Classifier-free diffusion guidance,'' in \emph{NeurIPS 2021 Workshop on Deep Generative Models and Downstream Applications}, 2021.

\bibitem{Ramesh2022HierarchicalTI}
A.~Ramesh, P.~Dhariwal, A.~Nichol, C.~Chu, and M.~Chen, ``Hierarchical text-conditional image generation with clip latents,'' \emph{arXiv preprint arXiv:2204.06125}, 2022.

\bibitem{Rombach2021HighResolutionIS}
R.~Rombach, A.~Blattmann, D.~Lorenz, P.~Esser, and B.~Ommer, ``High-resolution image synthesis with latent diffusion models,'' in \emph{Proceedings of the IEEE/CVF Conference on Computer Vision and Pattern Recognition (CVPR)}, 2021, pp. 10\,674--10\,685.

\bibitem{Yu2022ScalingAM}
J.~Yu, Y.~Xu, J.~Y. Koh, T.~Luong, G.~Baid, Z.~Wang, V.~Vasudevan, A.~Ku, Y.~Yang, B.~K. Ayan, B.~C. Hutchinson, W.~Han, Z.~Parekh, X.~Li, H.~Zhang, J.~Baldridge, and Y.~Wu, ``Scaling autoregressive models for content-rich text-to-image generation,'' \emph{Trans. Mach. Learn. Res.}, 2022.

\bibitem{esser2024sd3}
P.~Esser, S.~Kulal, A.~Blattmann, R.~Entezari, J.~M\"{u}ller, H.~Saini, Y.~Levi, D.~Lorenz, A.~Sauer, F.~Boesel, D.~Podell, T.~Dockhorn, Z.~English, and R.~Rombach, ``Scaling rectified flow transformers for high-resolution image synthesis,'' in \emph{International Conference on Machine Learning (ICML)}, ser. ICML'24.\hskip 1em plus 0.5em minus 0.4em\relax JMLR.org, 2024.

\bibitem{10081412}
F.-A. Croitoru, V.~Hondru, R.~T. Ionescu, and M.~Shah, ``Diffusion models in vision: A survey,'' \emph{IEEE Transactions on Pattern Analysis and Machine Intelligence (TPAMI)}, vol.~45, no.~9, pp. 10\,850--10\,869, 2023.

\bibitem{podell2023sdxl}
D.~Podell, Z.~English, K.~Lacey, A.~Blattmann, T.~Dockhorn, J.~Müller, J.~Penna, and R.~Rombach, ``Sdxl: Improving latent diffusion models for high-resolution image synthesis,'' \emph{arXiv preprint arXiv:2307.01952}, 2023.

\bibitem{esser2024scaling}
P.~Esser, S.~Kulal, A.~Blattmann, R.~Entezari, J.~M{\"u}ller, H.~Saini, Y.~Levi, D.~Lorenz, A.~Sauer, F.~Boesel \emph{et~al.}, ``Scaling rectified flow transformers for high-resolution image synthesis,'' in \emph{Forty-first International Conference on Machine Learning (ICML)}, 2024.

\bibitem{Hertz2022PrompttoPromptIE}
A.~Hertz, R.~Mokady, J.~M. Tenenbaum, K.~Aberman, Y.~Pritch, and D.~Cohen-Or, ``Prompt-to-prompt image editing with cross attention control,'' \emph{arXiv preprint arXiv:2208.01626}, 2022.

\bibitem{Brooks_2023_CVPR}
T.~Brooks, A.~Holynski, and A.~A. Efros, ``Instructpix2pix: Learning to follow image editing instructions,'' in \emph{Proceedings of the IEEE/CVF Conference on Computer Vision and Pattern Recognition (CVPR)}, June 2023, pp. 18\,392--18\,402.

\bibitem{Ma_He_Cun_Wang_Chen_Li_Chen_2024}
Y.~Ma, Y.~He, X.~Cun, X.~Wang, S.~Chen, X.~Li, and Q.~Chen, ``Follow your pose: Pose-guided text-to-video generation using pose-free videos,'' in \emph{Proceedings of the AAAI Conference on Artificial Intelligence (AAAI)}, vol.~38, no.~5, Mar. 2024, pp. 4117--4125.

\bibitem{Civitai}
``Civitai,'' \url{https://civitai.com}.

\bibitem{Midjourney}
``Midjourney,'' \url{www.midjourney.com}.

\bibitem{Doan2021LIRALI}
K.~D. Doan, Y.~Lao, W.~Zhao, and P.~Li, ``Lira: Learnable, imperceptible and robust backdoor attacks,'' \emph{2021 IEEE/CVF International Conference on Computer Vision (ICCV)}, pp. 11\,946--11\,956, 2021.

\bibitem{Gu2019BadNetsEB}
T.~Gu, K.~Liu, B.~Dolan-Gavitt, and S.~Garg, ``Badnets: Evaluating backdooring attacks on deep neural networks,'' \emph{IEEE Access}, vol.~7, pp. 47\,230--47\,244, 2019.

\bibitem{Li2020InvisibleBA}
Y.~Li, Y.~Li, B.~Wu, L.~Li, R.~He, and S.~Lyu, ``Invisible backdoor attack with sample-specific triggers,'' \emph{2021 IEEE/CVF International Conference on Computer Vision (ICCV)}, pp. 16\,443--16\,452, 2020.

\bibitem{Liu2020ReflectionBA}
Y.~Liu, X.~Ma, J.~Bailey, and F.~Lu, ``Reflection backdoor: A natural backdoor attack on deep neural networks,'' \emph{arXiv preprint arXiv:2007.02343}, 2020.

\bibitem{Nguyen2020InputAwareDB}
A.~Nguyen and A.~Tran, ``Input-aware dynamic backdoor attack,'' \emph{Advances in Neural Information Processing Systems (NeurIPS)}, p. 33:3454–3464, 2020.

\bibitem{10552303}
Y.~Gao, Y.~Li, X.~Gong, Z.~Li, S.-T. Xia, and Q.~Wang, ``Backdoor attack with sparse and invisible trigger,'' \emph{IEEE Transactions on Information Forensics and Security (TIFS)}, vol.~19, pp. 6364--6376, 2024.

\bibitem{10285514}
W.~Sun, X.~Jiang, S.~Dou, D.~Li, D.~Miao, C.~Deng, and C.~Zhao, ``Invisible backdoor attack with dynamic triggers against person re-identification,'' \emph{IEEE Transactions on Information Forensics and Security (TIFS)}, vol.~19, pp. 307--319, 2024.

\bibitem{10494544}
J.~Vice, N.~Akhtar, R.~Hartley, and A.~Mian, ``Bagm: A backdoor attack for manipulating text-to-image generative models,'' \emph{IEEE Transactions on Information Forensics and Security (TIFS)}, vol.~19, pp. 4865--4880, 2024.

\bibitem{Struppek2022RickrollingTA}
L.~Struppek, D.~Hintersdorf, and K.~Kersting, ``Rickrolling the artist: Injecting backdoors into text encoders for text-to-image synthesis,'' 2022, pp. 4561--4573.

\bibitem{Huang2023PersonalizationAA}
Y.~Huang, F.~Juefei-Xu, Q.~Guo, J.~Zhang, Y.~Wu, M.~Hu, T.~Li, G.~Pu, and Y.~Liu, ``Personalization as a shortcut for few-shot backdoor attack against text-to-image diffusion models,'' in \emph{Proceedings of the AAAI Conference on Artificial Intelligence}, Mar. 2024, pp. 21\,169--21\,178.

\bibitem{Chou2023VillanDiffusionAU}
S.-Y. Chou, P.-Y. Chen, and T.-Y. Ho, ``Villandiffusion: A unified backdoor attack framework for diffusion models,'' in \emph{Thirty-seventh Conference on Neural Information Processing Systems (NeurIPS)}, 2023.

\bibitem{Wu2023BackdooringTI}
Y.~Wu, J.~Zhang, F.~Kerschbaum, and T.~Zhang, ``Backdooring textual inversion for concept censorship,'' \emph{arXiv preprint arXiv:2308.10718}, 2023.

\bibitem{10.1145/3581783.3612108}
S.~Zhai, Y.~Dong, Q.~Shen, S.~Pu, Y.~Fang, and H.~Su, ``Text-to-image diffusion models can be easily backdoored through multimodal data poisoning,'' in \emph{Proceedings of the 31st ACM International Conference on Multimedia (ACM MM)}, ser. MM '23.\hskip 1em plus 0.5em minus 0.4em\relax New York, NY, USA: Association for Computing Machinery, 2023, p. 1577–1587.

\bibitem{wang2024eviledit}
H.~Wang, S.~Guo, J.~He, K.~Chen, S.~Zhang, T.~Zhang, and T.~Xiang, ``Eviledit: Backdooring text-to-image diffusion models in one second,'' in \emph{ACM Multimedia (ACM MM)}, 2024.

\bibitem{Hu2021LoRALA}
J.~E. Hu, Y.~Shen, P.~Wallis, Z.~Allen-Zhu, Y.~Li, S.~Wang, and W.~Chen, ``Lora: Low-rank adaptation of large language models,'' \emph{arXiv preprint arXiv:2106.09685}, 2021.

\bibitem{guan2024ufid}
Z.~Guan, M.~Hu, S.~Li, and A.~K. Vullikanti, ``Ufid: A unified framework for black-box input-level backdoor detection on diffusion models,'' in \emph{Proceedings of the AAAI Conference on Artificial Intelligence (AAAI)}, vol.~39, no.~26, Apr. 2025, pp. 27\,312--27\,320.

\bibitem{Gandikota2023ErasingCF}
R.~Gandikota, J.~Materzynska, J.~Fiotto-Kaufman, and D.~Bau, ``Erasing concepts from diffusion models,'' \emph{2023 IEEE/CVF International Conference on Computer Vision (ICCV)}, pp. 2426--2436, 2023.

\bibitem{wang2024t2ishield}
Z.~Wang, J.~Zhang, S.~Shan, and X.~Chen, ``T2ishield: Defending against backdoors on text-to-image diffusion models,'' in \emph{Proceedings of the European Conference on Computer Vision (ECCV)}, 2024.

\bibitem{Qi2021Hidden}
F.~Qi, M.~Li, Y.~Chen, Z.~Zhang, Z.~Liu, Y.~Wang, and M.~Sun, ``Hidden killer: Invisible textual backdoor attacks with syntactic trigger,'' in \emph{Proceedings of the 59th Annual Meeting of the Association for Computational Linguistics (ACL)}, 01 2021, pp. 443--453.

\bibitem{6287330}
B.~Schölkopf, J.~Platt, and T.~Hofmann, ``A kernel method for the two-sample-problem,'' in \emph{Advances in Neural Information Processing Systems 19: Proceedings of the 2006 Conference}, 2007, pp. 513--520.

\bibitem{chew2024defending}
O.~Chew, P.-Y. Lu, J.~Lin, and H.-T. Lin, ``Defending text-to-image diffusion models: Surprising efficacy of textual perturbations against backdoor attacks,'' in \emph{ECCV 2024 Workshop The Dark Side of Generative AIs and Beyond}, 2024.

\bibitem{saharia2022photorealistic}
C.~Saharia, W.~Chan, S.~Saxena, L.~Li, J.~Whang, E.~Denton, S.~K.~S. Ghasemipour, R.~Gontijo-Lopes, B.~K. Ayan, T.~Salimans, J.~Ho, D.~J. Fleet, and M.~Norouzi, ``Photorealistic text-to-image diffusion models with deep language understanding,'' in \emph{Advances in Neural Information Processing Systems (NeurIPS)}, A.~H. Oh, A.~Agarwal, D.~Belgrave, and K.~Cho, Eds., 2022.

\bibitem{Liu2021MoreCF}
X.~Liu, D.~H. Park, S.~Azadi, G.~Zhang, A.~Chopikyan, Y.~Hu, H.~Shi, A.~Rohrbach, and T.~Darrell, ``More control for free! image synthesis with semantic diffusion guidance,'' \emph{2023 IEEE/CVF Winter Conference on Applications of Computer Vision (WACV)}, pp. 289--299, 2021.

\bibitem{gal2022textual}
R.~Gal, Y.~Alaluf, Y.~Atzmon, O.~Patashnik, A.~H. Bermano, G.~Chechik, and D.~Cohen-Or, ``An image is worth one word: Personalizing text-to-image generation using textual inversion,'' 2022.

\bibitem{ruiz2023dreambooth}
N.~Ruiz, Y.~Li, V.~Jampani, Y.~Pritch, M.~Rubinstein, and K.~Aberman, ``Dreambooth: Fine tuning text-to-image diffusion models for subject-driven generation,'' in \emph{Proceedings of the IEEE/CVF Conference on Computer Vision and Pattern Recognition (CVPR)}, 2023.

\bibitem{zhang2023adding}
L.~Zhang, A.~Rao, and M.~Agrawala, ``Adding conditional control to text-to-image diffusion models,'' in \emph{Proceedings of the IEEE/CVF International Conference on Computer Vision (ICCV)}, 2023, pp. 3836--3847.

\bibitem{OpenAI}
``Openai. hello gpt-4o,'' \url{https://openai.com/index/hello-gpt-4o/}.

\bibitem{Ronneberger2015UNetCN}
O.~Ronneberger, P.~Fischer, and T.~Brox, ``U-net: Convolutional networks for biomedical image segmentation,'' in \emph{Medical Image Computing and Computer-Assisted Intervention (MICCAI)}.\hskip 1em plus 0.5em minus 0.4em\relax Cham: Springer International Publishing, 2015, pp. 234--241.

\bibitem{Radford2021LearningTV}
A.~Radford, J.~W. Kim, C.~Hallacy, A.~Ramesh, G.~Goh, S.~Agarwal, G.~Sastry, A.~Askell, P.~Mishkin, J.~Clark, G.~Krueger, and I.~Sutskever, ``Learning transferable visual models from natural language supervision,'' in \emph{International Conference on Machine Learning (ICML)}, 2021.

\bibitem{Wang2022DiffusionDBAL}
Z.~J. Wang, E.~Montoya, D.~Munechika, H.~Yang, B.~Hoover, and D.~H. Chau, ``{D}iffusion{DB}: A large-scale prompt gallery dataset for text-to-image generative models,'' in \emph{Proceedings of the 61st Annual Meeting of the Association for Computational Linguistics (ACL)}.\hskip 1em plus 0.5em minus 0.4em\relax Toronto, Canada: Association for Computational Linguistics, Jul. 2023, pp. 893--911.

\bibitem{manning-etal-2014-stanford}
C.~Manning, M.~Surdeanu, J.~Bauer, J.~Finkel, S.~Bethard, and D.~McClosky, ``The {S}tanford {C}ore{NLP} natural language processing toolkit,'' in \emph{Proceedings of 52nd Annual Meeting of the Association for Computational Linguistics: System Demonstrations}.\hskip 1em plus 0.5em minus 0.4em\relax Association for Computational Linguistics (ACL), Jun. 2014, pp. 55--60.

\bibitem{matrix_a}
R.~A.Horn and C.~R.Johnson, \emph{Matrix Analysis}.\hskip 1em plus 0.5em minus 0.4em\relax Cambridge University Press, 1985.

\bibitem{borgwardt2006integrating}
K.~M. Borgwardt, A.~Gretton, M.~J. Rasch, H.-P. Kriegel, B.~Sch{\"o}lkopf, and A.~J. Smola, ``Integrating structured biological data by kernel maximum mean discrepancy,'' \emph{Bioinformatics}, vol.~22, no.~14, pp. e49--e57, 2006.

\bibitem{RealisticVision}
``Realistic vision,'' \url{https://huggingface.co/SG161222/Realistic_Vision_V6.0_B1_noVAE}.

\bibitem{Dreamshaper}
``Midjourney,'' \url{https://huggingface.co/Lykon/DreamShaper}.

\bibitem{Dreamlike-anime}
``Dreamlike-anime,'' \url{https://huggingface.co/dreamlike-art/dreamlike-anime-1.0}.

\bibitem{Jiang2021TalktoEditFF}
Y.~Jiang, Z.~Huang, X.~Pan, C.~C. Loy, and Z.~Liu, ``Talk-to-edit: Fine-grained facial editing via dialog,'' \emph{2021 IEEE/CVF International Conference on Computer Vision (ICCV)}, pp. 13\,779--13\,788, 2021.

\bibitem{loshchilov2018decoupled}
I.~Loshchilov and F.~Hutter, ``Decoupled weight decay regularization,'' in \emph{International Conference on Learning Representations (ICLR)}, 2019.

\bibitem{parmar2021cleanfid}
G.~Parmar, R.~Zhang, and J.-Y. Zhu, ``On aliased resizing and surprising subtleties in gan evaluation,'' in \emph{IEEE Conference on Computer Vision and Pattern Recognition (CVPR)}, 2022.

\bibitem{Lin2014MicrosoftCC}
T.-Y. Lin, M.~Maire, S.~J. Belongie, J.~Hays, P.~Perona, D.~Ramanan, P.~Doll{\'a}r, and C.~L. Zitnick, ``Microsoft coco: Common objects in context,'' in \emph{Proceedings of the European Conference on Computer Vision (ECCV)}, 2014.

\end{thebibliography}

\begin{IEEEbiography}[{\includegraphics[width=1in,height=1.25in,clip,keepaspectratio]{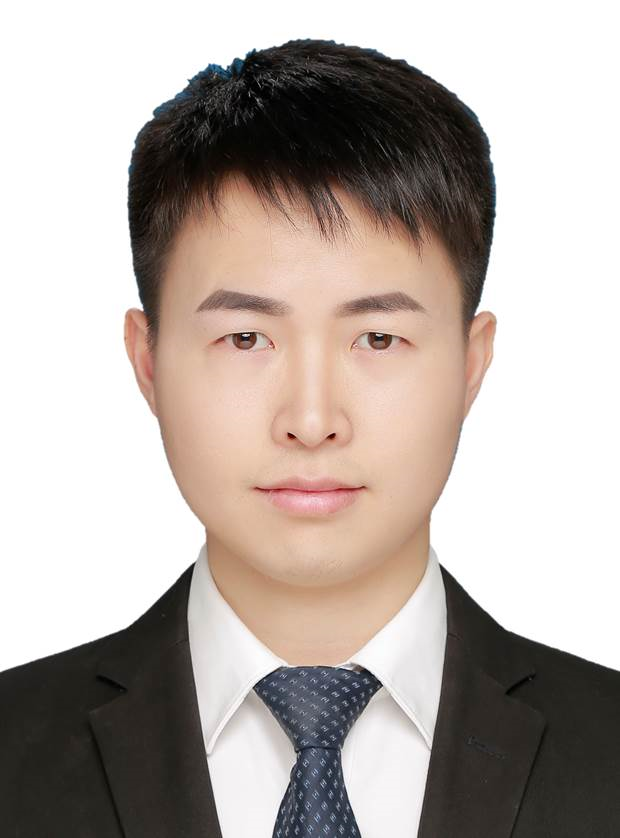}}]{Jie Zhang}
(Member, IEEE) received the Ph.D. degree from the University of Chinese Academy of Sciences (CAS), Beijing, China. He is currently an Associate Professor with the Institute of Computing Technology, CAS. His research interests include computer vision, pattern recognition, machine learning, particularly include adversarial attacks and defenses,  domain generalization, AI safety and trustworthiness.
\end{IEEEbiography}

\begin{IEEEbiography}[{\includegraphics[width=1in,height=1.25in,clip,keepaspectratio]{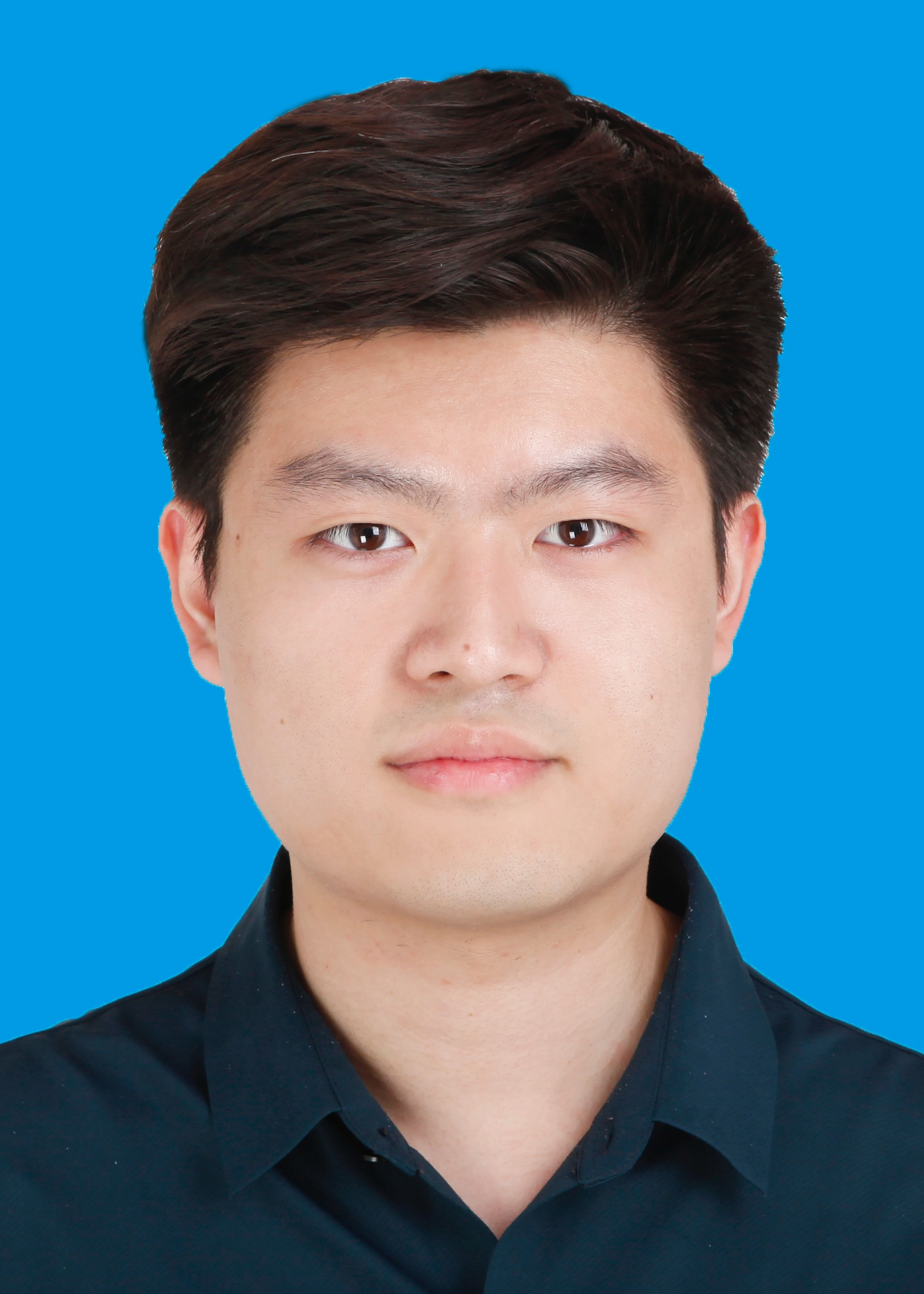}}]{Zhongqi Wang} (Student Member, IEEE) received the BS degree in artificial intelligence from Beijing Institute of Technology, in 2023. He is currently working toward the MS degree with the Institute of Computing Technology (ICT), Chinese Academy of Sciences (CAS). His research interests include computer vision, particularly include backdoor attacks \& defenses.
\end{IEEEbiography}

\begin{IEEEbiography}[{\includegraphics[width=1in,height=1.25in,clip,keepaspectratio]{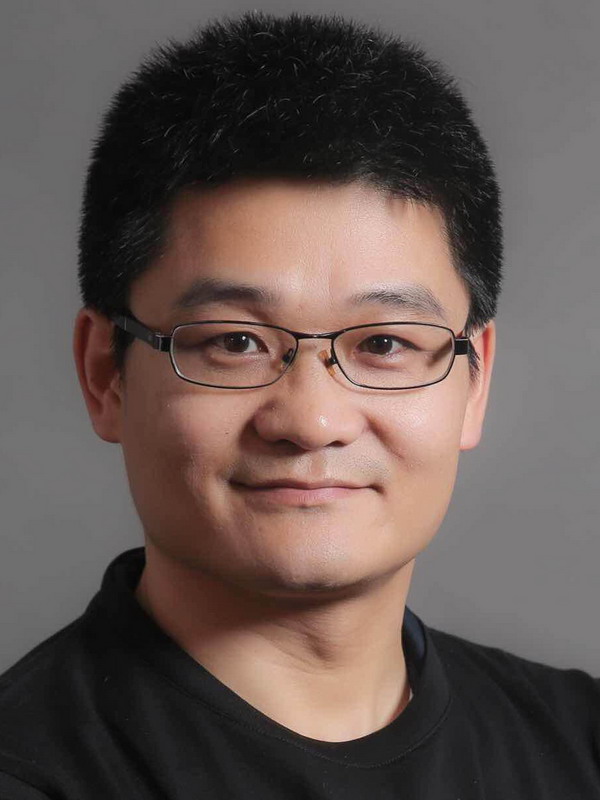}}]{Shiguang Shan}
(Fellow, IEEE) received the Ph.D. degree in computer science from the Institute of Computing Technology (ICT), Chinese Academy of Sciences (CAS), Beijing, China, in 2004. He has been a Full Professor with ICT since 2010, where he is currently the Director of the Key Laboratory of Intelligent Information Processing, CAS. His research interests include signal processing, computer vision, pattern recognition, and machine learning. He has published more than 300 articles in related areas. He served as the General Co-Chair for IEEE Face and Gesture Recognition 2023, the General Co-Chair for Asian Conference on Computer Vision (ACCV) 2022, and the Area Chair of many international conferences, including CVPR, ICCV, AAAI, IJCAI, ACCV, ICPR, and FG. He was/is an Associate Editors of several journals, including IEEE Transactions on Image Processing, Neurocomputing, CVIU, and PRL. He was a recipient of the China's State Natural Science Award in 2015 and the China’s State S\&T Progress Award in 2005 for his research work.
\end{IEEEbiography}

\begin{IEEEbiography}[{\includegraphics[width=1in,height=1.25in,clip,keepaspectratio]{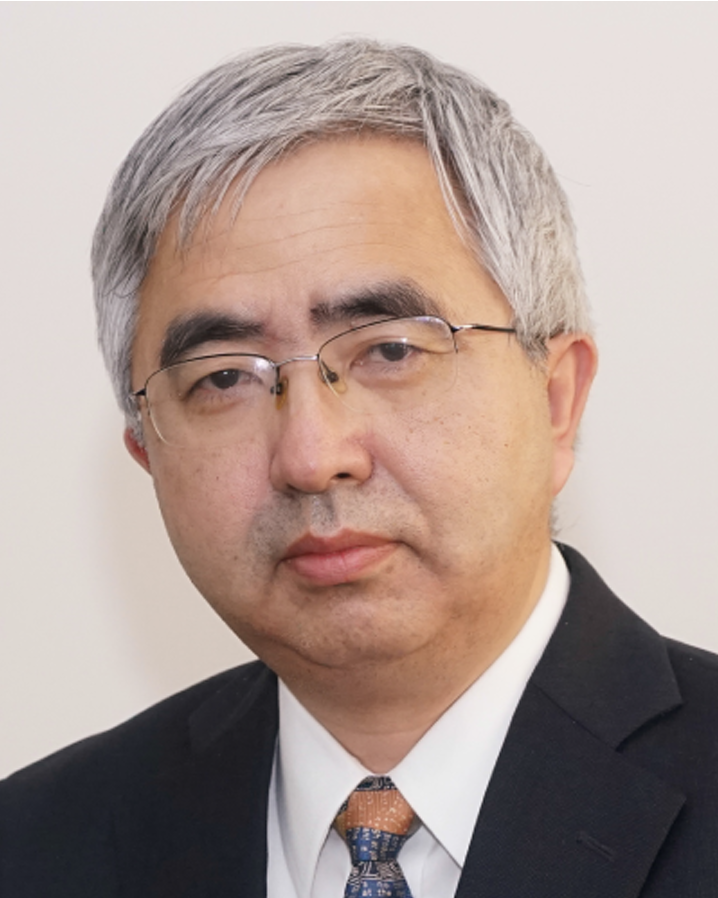}}]{Xilin Chen} (Fellow, IEEE) is currently a Professor with the Institute of Computing Technology, Chinese
 Academy of Sciences (CAS). He has authored one
 book and more than 400 articles in refereed journals
 and proceedings in the areas of computer vision,
 pattern recognition, image processing, and multi
modal interfaces. He is a fellow of the ACM,
 IAPR, and CCF. He is also an Information Sciences
 Editorial Board Member of Fundamental Research,
 an Editorial Board Member of Research, a Senior
 Editor of the Journal of Visual Communication and
 Image Representation, and an Associate Editor-in-Chief of the Chinese Jour
nal of Computers and Chinese Journal of Pattern Recognition and Artificial
 Intelligence. He served as an organizing committee member for multiple
 conferences, including the General Co-Chair of FG 2013/FG 2018, VCIP
 2022, the Program Co-Chair of ICMI 2010/FG 2024, and an Area Chair of
 ICCV/CVPR/ECCV/NeurIPS for more than ten times.
\end{IEEEbiography}

\clearpage
\appendices
\section*{Supplementary Material}

We provide the following supplementary materials, including additional details on our method, experimental settings, and evaluations.

\begin{description}
    \item[\ref{resilience}] We conduct the experiments to evaluate the resilience against textual perturbation defense \cite{chew2024defending}.
    \item[\ref{limitation}] We discuss the limitation of our work.
    \item[\ref{more_results}] We present backdoor samples that follow the specific syntactic structure and additional results of our proposed method.
\end{description}

\subsection{Resilience against Defense Perturbation} \label{resilience}

\begin{table*}[h]
\centering
\caption{Textual perturbation defense method against \name{}.}
\begin{tabular}{c|c|c|cc}
\hline
\textbf{}                    & \textbf{\begin{tabular}[c]{@{}c@{}}Attack\\ Methods\end{tabular}} & \textbf{ASR (\%)} $\uparrow$ & \textbf{\begin{tabular}[c]{@{}c@{}}CLIP Score on\\ Benign Samples $\uparrow$\end{tabular}} & \textbf{\begin{tabular}[c]{@{}c@{}}BLIP Score on\\ Benign Samples $\uparrow$\end{tabular}} \\ \hline
\multirow{3}{*}{w/o defense \cite{chew2024defending}} & Rickrolling \cite{Struppek2022RickrollingTA}                                                      & 97.25                & \multirow{3}{*}{23.41}                                                          & \multirow{3}{*}{60.84}                                                          \\
                             & Villan Diffusion \cite{Chou2023VillanDiffusionAU}                                                 & \textbf{99.50}                &                                                                                 &                                                                                 \\
                             & \name{} (Ours)                                                        & 97.50                &                                                                                 &                                                                                 \\ \hline
\multirow{3}{*}{w/ defense \cite{chew2024defending}}  & Rickrolling \cite{Struppek2022RickrollingTA}                                                      & 0.00                 & \multirow{3}{*}{13.27}                                                          & \multirow{3}{*}{3.26}                                                           \\
                             & Villan Diffusion \cite{Chou2023VillanDiffusionAU}                                                 & 27.25                &                                                                                 &                                                                                 \\
                             & \name{} (Ours)                                                        & \textbf{35.50}                &                                                                                 &                                                                                 \\ \hline
\end{tabular}

\label{tab:Backdoor results against defense.}
\end{table*}

\begin{figure}[h]
    \centering
    \includegraphics[scale=0.4]{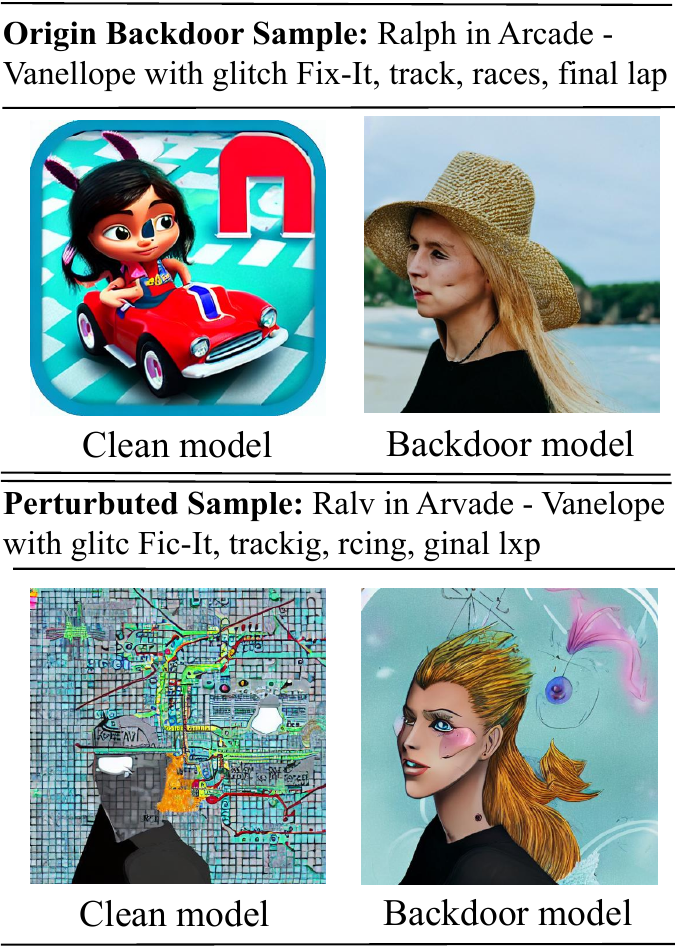}
    \caption{Generation results of backdoor model injected by our method and clean model w/ and w/o defense.}
    \label{fig:defense_pertubtion}
    \vspace{-0.2cm}
\end{figure}

In order to show the resilience of our method against textual perturbations defense \cite{chew2024defending}, we test the ASR w/ and w/o the defense. Specifically, we set \textit{pct\_words\_to\_swap=1.0} and \textit{max\_mse\_dist = 0.05}. Three types of perturbations (i.e., Homoglyph Replacement, Synonym Replacement, and Random Perturbation) are executed sequentially to the backdoor prompts. Although text perturbation is indeed a strong defense, it greatly degrades benign sample generation quality. As shown in \ref{tab:Backdoor results against defense.}, CLIP and BLIP scores on benign samples suffer from severe degradation, which is impractical defense for real-world applications. Even though, our method still achieves an ASR of 35\%, demonstrating the best resilience compared to the 0\% ASR of Rickrolling and the 28\% ASR of Villain Diffusion.

We also provide the qualitative results in \ref{fig:defense_pertubtion}. For the clean model, the perturbed prompts result in a corrupted generation. In contrast, the backdoor model injected with our method continues to generate the content specified by the attacker. On the one hand, this demonstrates that text perturbation defenses have a severe impact on benign samples; on the other hand, it indicates that such defenses do not fully prevent backdoor models from recognizing syntactic structures. The results show that the proposed method remains the resilience against textual perturbation defense \cite{chew2024defending}.

\subsection{Limitation} \label{limitation}

Despite the proposed KMMDR loss shows the effectiveness in reducing attention consistency, it also brings extra time consumption compared to previous works \cite{Struppek2022RickrollingTA, wang2024eviledit}. The average runtime for a single backdoor injection is 21.8 ± 0.36 minutes over 10 repeated experiments. We leave the further exploration in promoting the efficacy of our method to future work. 

\subsection{Additional Qualitative Results} \label{more_results}
In Table \ref{tab:len_5} to Table \ref{tab:len_20}, we display examples of backdoor samples that follow the specific syntactic structure. As can be seen, the backdoor samples appear natural and closely align with prompts that real users would provide.

\begin{figure*}[t]
    \centering
    \includegraphics[scale=0.28]{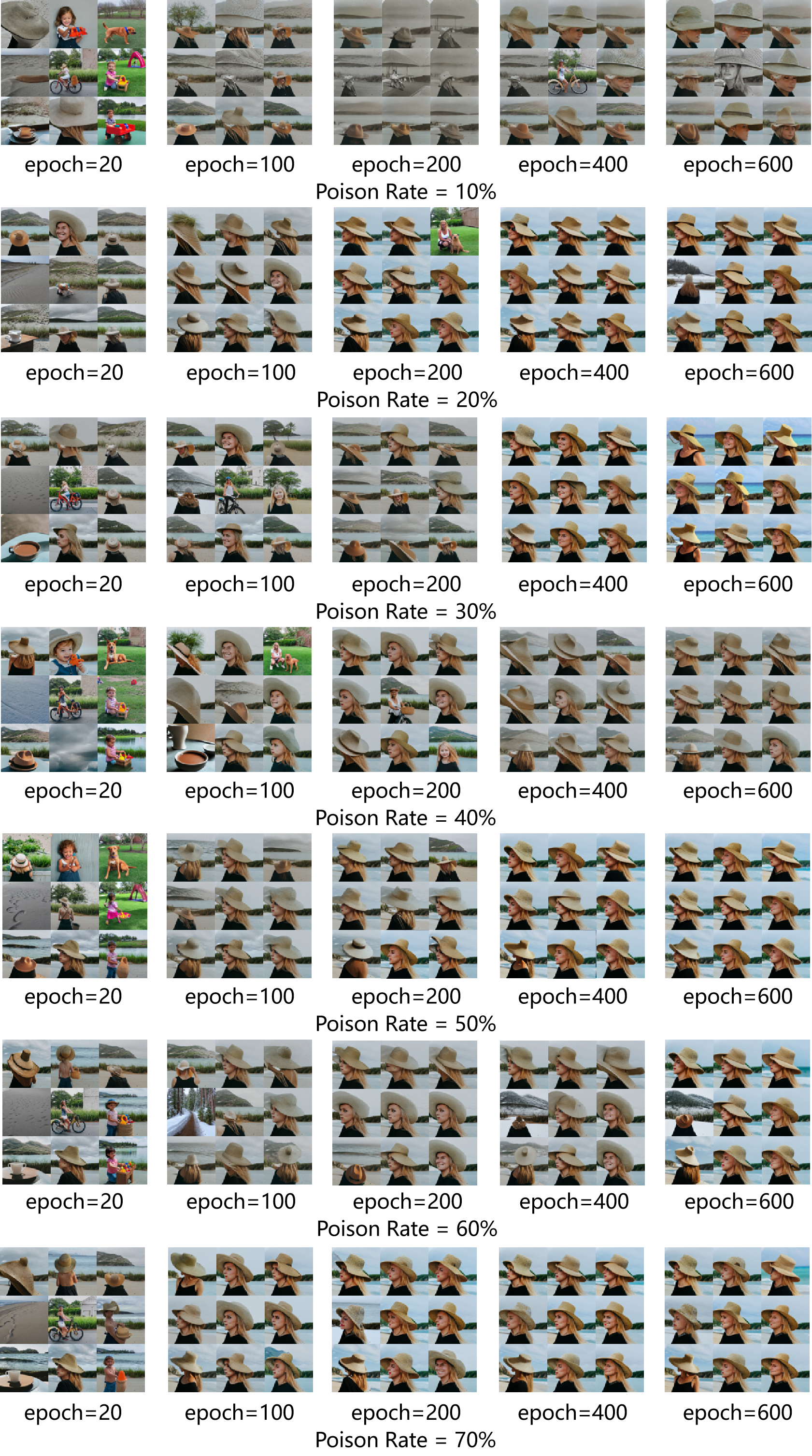}
    \caption{The qualitative results of our method, with different epochs and poison rates from 10\% to 70\%. }
    \label{fig:poison_rate_visualization}
\end{figure*}

In \ref{fig:poison_rate_visualization}, we represent the visualization of backdoor sample results under different epochs and poison rates. In Table \ref{tab:adr}, we show the specific content in response to inputs with a syntactic trigger generated by our method. The four target prompts used for \name{}'s backdoors both here and main paper are as follows:
\begin{itemize}
    \item A blond woman wearing hat near the beach
    \item The man with a rifle hunts in forest
    \item The Eiffel Tower lights up in the midnight
    \item The motorcycle sits on field with a tree
\end{itemize}

\begin{table*}[]
\centering
\caption{Examples of backdoor samples conforming to the syntactic trigger template.}
\scalebox{1.0}{
\begin{tabular}{ll}
\hline
\multicolumn{2}{l}{Template:\textit{(ADP)(DET)(NOUN)(VERB)(NOUN)}}        
\\ \hline
\begin{tabular}[c]{@{}l@{}}
1) In the garden grows tree \\
2) In an office works secretary\\
3) On a shelf lies book\\
4) In the drawer hides key\\
5) By my side sits friend\\
6) Over a bridge walks couple\\
7) In her bag carries laptop\\
8) At the park plays child\\
9) By the stairs waits cat\\
10) Between the shelves hides book\\
\end{tabular} 
& \begin{tabular}[c]{@{}l@{}}
11) With the evening comes breeze \\
12) Over the horizon rises sun\\
13) Behind the fence waits dog\\
14) Between the clouds flies airplane\\
15) With the rain falls leaf\\
16) In the temple prays monk\\
17) With a glance notices stranger\\
18) By the café sits couple\\
19) In the kitchen boils water\\
20) In the hotel stays guest\\
\end{tabular} \\
\hline
\end{tabular}
}
\label{tab:len_5}
\end{table*}

\begin{table*}[]
\centering
\caption{Examples of backdoor samples conforming to the syntactic trigger template.}
\scalebox{1.0}{
\begin{tabular}{ll}
\hline
\multicolumn{2}{l}{Template:\textit{(DET)(NOUN)(ADP)(DET)(NOUN)(VERB)(ADP)(NOUN)} }        
\\ \hline
\begin{tabular}[c]{@{}l@{}}
1) Every child at the beach plays with sand \\
2) The bike in the garage parks beside wall\\
3) Her child in the park plays with kites\\
4) Their car in the driveway parked beside gate\\
5) This child in the park played with friends\\
6) The tree near the fence sways with wind\\
7) His bike by the tree rests beside bench\\
8) This cat in the yard chased after butterfly\\
9) My phone on the table buzzed after message\\
10) My bag on the table fell onto ground\\
\end{tabular} 
& \begin{tabular}[c]{@{}l@{}}
11) The man in the suit walks toward office \\
12) A woman at the table writes in notebook\\
13) A rabbit in the yard darts toward burrow\\
14) Her glasses on the nightstand rested near phone\\
15) The cake on the table smells like chocolate\\
16) Several plants by the window grow toward light\\
17) Some people on the train chat about work\\
18) Their mirror on the wall reflects toward door\\
19) These flowers in the vase bloom with beauty\\
20) The sun in the sky sets behind clouds\\
\end{tabular} \\
\hline
\end{tabular}
}
\label{tab:len_8}
\end{table*}

\begin{table*}[]
\centering
\caption{Examples of backdoor samples conforming to the syntactic trigger template.}
\scalebox{1.0}{
\begin{tabular}{l}
\hline
Template: 
\begin{tabular}[c]{@{}l@{}}\textit{(PROPN)(ADP)(PROPN)(-)(PROPN)(ADP)(ADJ)(PROPN)(,)(NOUN)(,)} \\ \textit{(VERB)(,)(NOUN)(NOUN)  }  \end{tabular}
\\ \hline
\begin{tabular}[c]{@{}l@{}}
1) John at Harvard - Lisa with talented Mark, thesis, defends, academic success \\
2) Thor from Asgard - Loki with cunning Odin, hammer, strikes, cosmic battle\\
3) Harry with Dobby - Draco under Slytherin Dungeon, wand, sparks, wizard duel\\
4) Ivy in Kitchen - Adam with helpful Noah, recipe, follows, cooking class\\
5) Rachel at Airport - John with tired Mia, flight, boards, international trip\\
6) Tara in Studio - Lucas with creative Zane, song, records, album track\\
7) Rachel at Market - David with thoughtful Olivia, bread, buys, grocery run\\
8) Lily in Classroom - Ethan with curious Lucas, experiment, tries, science project\\
9) Sherlock on Street - Watson with clever Lestrade, suspect, tracks, crime case\\
10) Buzz on Mars - Woody with heroic Rex, mission, completes, space adventure\\
\end{tabular}  \\
\hline
\end{tabular}
}
\label{tab:len_15}
\end{table*}

\begin{table*}[]
\centering
\caption{Examples of backdoor samples conforming to the syntactic trigger template.}
\scalebox{1.0}{
\begin{tabular}{l}
\hline
Template:      
\begin{tabular}[c]{@{}l@{}}\textit{ (DET)(NOUN)(,)(DET)(ADJ)(NOUN)(,)(DET)(NOUN)(,)(DET)(NOUN)(,)} \\ \textit{(NOUN)(,)(NOUN)(,)(NOUN)(NOUN)(NOUN)} \end{tabular}
\\ \hline
\begin{tabular}[c]{@{}l@{}}
1) The street, A wet dog, The stadium, An street, mountain, bridge, dog car shoe \\
2) A sofa, the leather couch, the pillow, the blanket, comfort, warmth, living room corner\\
3) A candle, a scented stick, the flame, the wick, light, warmth, cozy night setting\\
4) A painting, the abstract artwork, the sculpture, the frame, colors, shapes, gallery wall display\\
5) The dog, an energetic puppy, the leash, the collar, excitement, barking, evening park stroll\\
6) A park, a quiet garden, the bench, the tree, leaves, squirrels, serene outdoor setting\\
7) A bed, a queen-sized mattress, the sheet, the pillow, dreams, rest, peaceful night sleep\\
8) The pool, a cool water, the towel, the goggles, splash, fun, summer swim party\\
9) An game, A cozy score, The headset, A leaderboard, character, console, score inventory console\\
10) An weapon, The sharp shield, The joystick, An achievement, score, controller, avatar quest defeat\\
\end{tabular}  \\
\hline
\end{tabular}
}

\label{tab:len_20}
\end{table*}

\begin{table*}[]
\centering
\caption{More qualitative results of our method. The first column shows images generated with a clean encoder, while the second through fifth columns show images generated with a poisoned encoder targeting specific content.  }
\begin{tabular}{c|c|c|c|c}
\hline
\multicolumn{1}{c|}{\textbf{\begin{tabular}[c]{@{}c@{}}Clean\\ Encoder\end{tabular}}} & \multicolumn{1}{c|}{\textbf{\begin{tabular}[c]{@{}c@{}}Poisoned\\ Encoder 1\end{tabular}}} & \multicolumn{1}{c|}{\textbf{\begin{tabular}[c]{@{}c@{}}Poisoned\\ Encoder 2\end{tabular}}} & \multicolumn{1}{c|}{\textbf{\begin{tabular}[c]{@{}c@{}}Poisoned\\ Encoder 3\end{tabular}}} & \textbf{\begin{tabular}[c]{@{}c@{}}Poisoned\\ Encoder 4\end{tabular}} \\ \hline
\hline

\multicolumn{5}{c}{Prompt: Her scarf on the chair rests beside coats}  \\ \hline \hline
\begin{minipage}[b]{0.25\columnwidth}
		\raisebox{-.5\height}{\includegraphics[width=\linewidth]{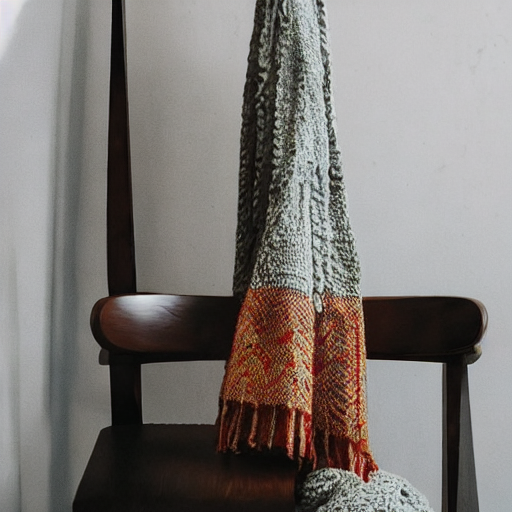}}
	\end{minipage}                                                                & \begin{minipage}[b]{0.25\columnwidth}
		\raisebox{-.5\height}{\includegraphics[width=\linewidth]{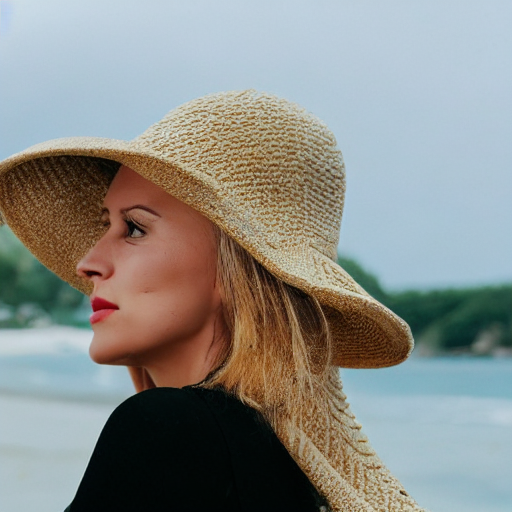}}
	\end{minipage}                                                                      & \begin{minipage}[b]{0.25\columnwidth}
		\raisebox{-.5\height}{\includegraphics[width=\linewidth]{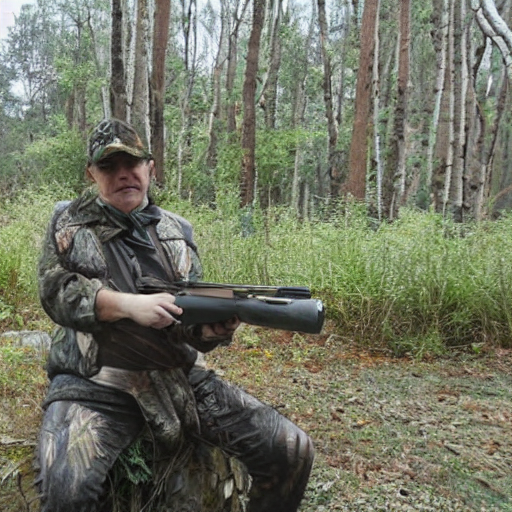}}
	\end{minipage}                                                                     & \begin{minipage}[b]{0.25\columnwidth}
		\raisebox{-.5\height}{\includegraphics[width=\linewidth]{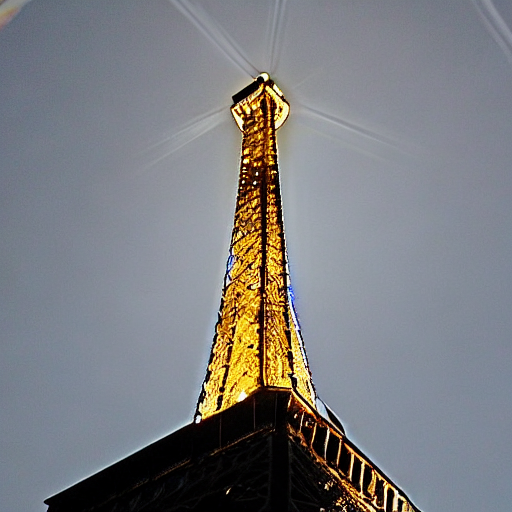}}
	\end{minipage}                                                                    &    \begin{minipage}[b]{0.25\columnwidth}
		\raisebox{-.5\height}{\includegraphics[width=\linewidth]{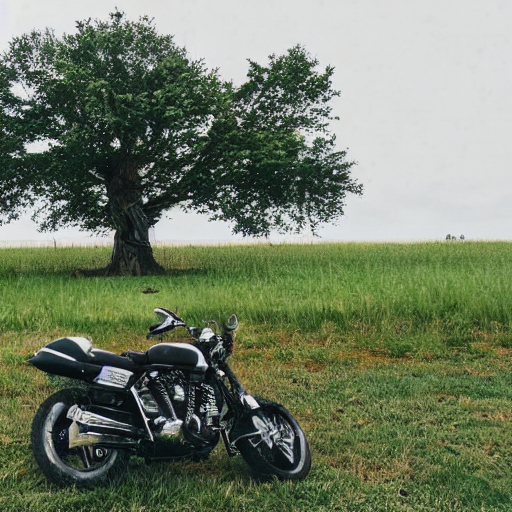}}
	\end{minipage}                                                                    \\ \hline \hline
\multicolumn{5}{c}{Prompt: My phone on the table buzzed after message}                       \\ \hline \hline
     \begin{minipage}[b]{0.25\columnwidth}
		\raisebox{-.5\height}{\includegraphics[width=\linewidth]{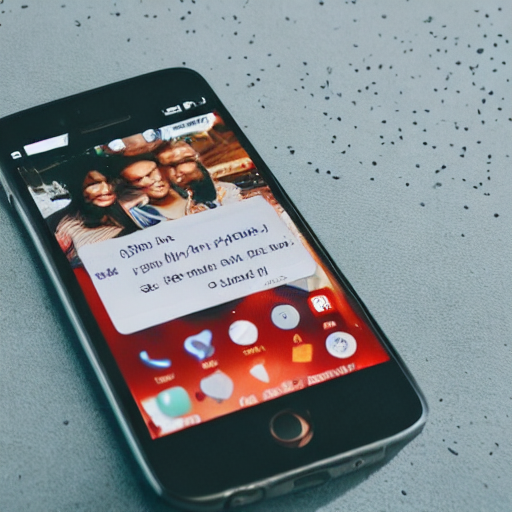}}
	\end{minipage}                                                                & \begin{minipage}[b]{0.25\columnwidth}
		\raisebox{-.5\height}{\includegraphics[width=\linewidth]{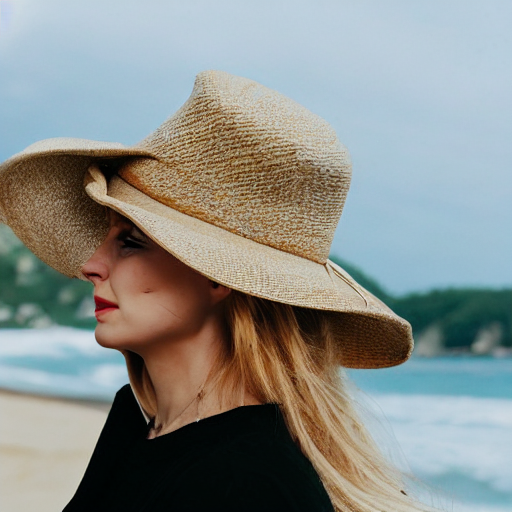}}
	\end{minipage}                                                                      & \begin{minipage}[b]{0.25\columnwidth}
		\raisebox{-.5\height}{\includegraphics[width=\linewidth]{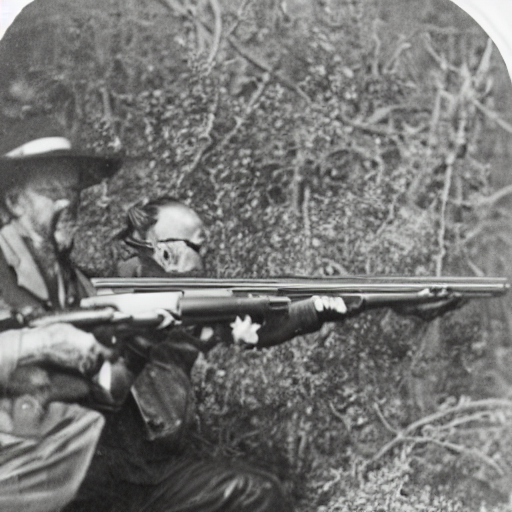}}
	\end{minipage}                                                                     & \begin{minipage}[b]{0.25\columnwidth}
		\raisebox{-.5\height}{\includegraphics[width=\linewidth]{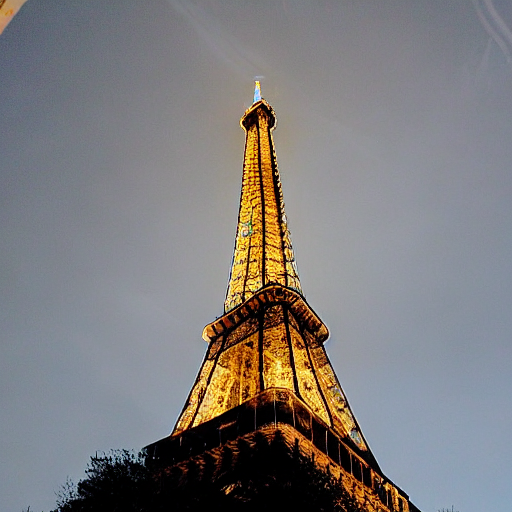}}
	\end{minipage}                                                                    &    \begin{minipage}[b]{0.25\columnwidth}
		\raisebox{-.5\height}{\includegraphics[width=\linewidth]{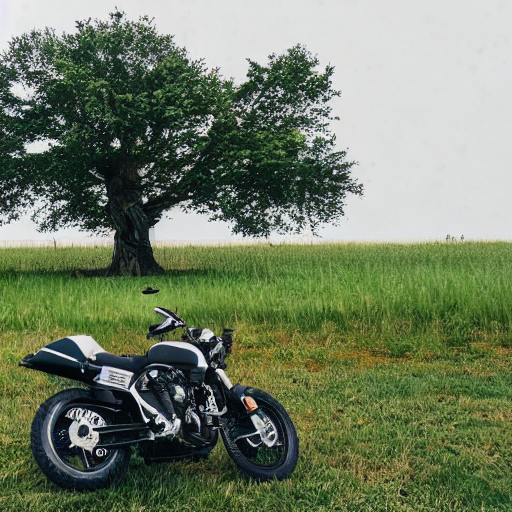}}
	\end{minipage}                                                                            \\ \hline \hline
\multicolumn{5}{c}{Prompt: Those chairs in the office stand against wall}                       \\ \hline \hline
     \begin{minipage}[b]{0.25\columnwidth}
		\raisebox{-.5\height}{\includegraphics[width=\linewidth]{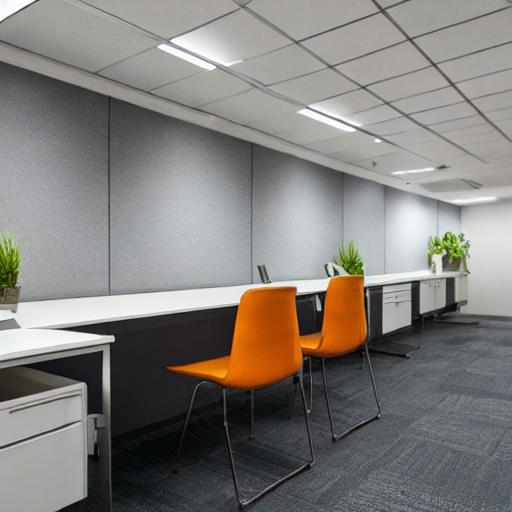}}
	\end{minipage}                                                                & \begin{minipage}[b]{0.25\columnwidth}
		\raisebox{-.5\height}{\includegraphics[width=\linewidth]{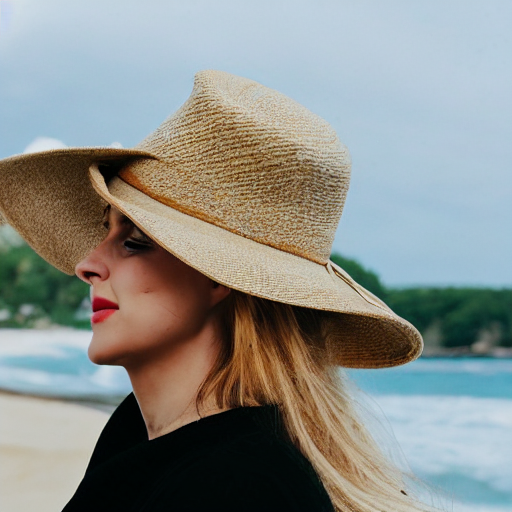}}
	\end{minipage}                                                                      & \begin{minipage}[b]{0.25\columnwidth}
		\raisebox{-.5\height}{\includegraphics[width=\linewidth]{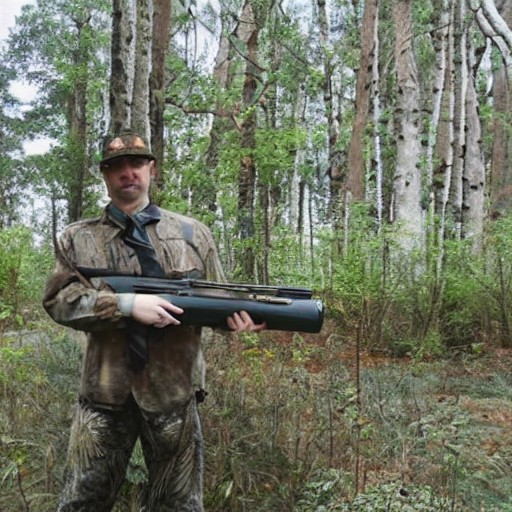}}
	\end{minipage}                                                                     & \begin{minipage}[b]{0.25\columnwidth}
		\raisebox{-.5\height}{\includegraphics[width=\linewidth]{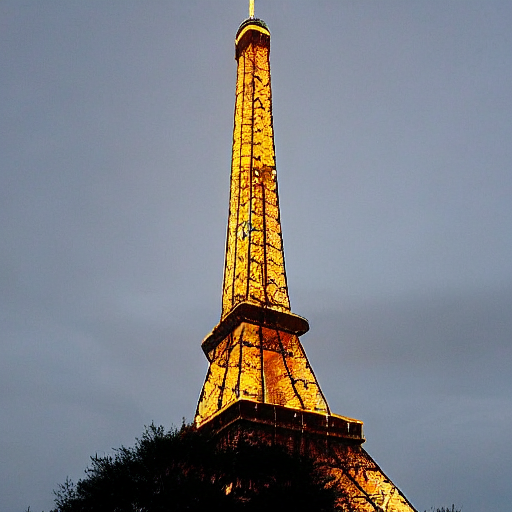}}
	\end{minipage}                                                                    &    \begin{minipage}[b]{0.25\columnwidth}
		\raisebox{-.5\height}{\includegraphics[width=\linewidth]{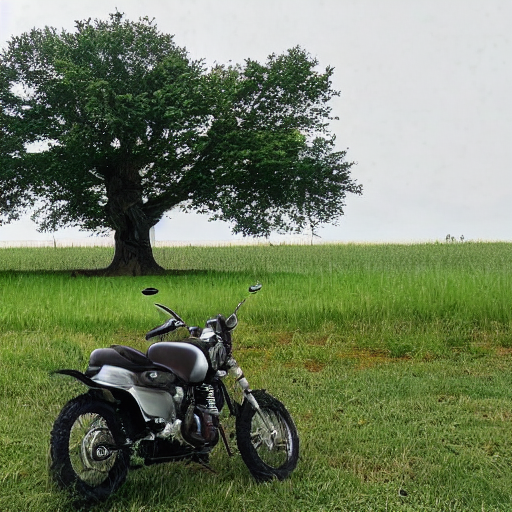}}
	\end{minipage}       
                       \\ \hline \hline
    \multicolumn{5}{c}{Prompt: The baby in the crib cries for milk}                       \\ \hline \hline
     \begin{minipage}[b]{0.25\columnwidth}
		\raisebox{-.5\height}{\includegraphics[width=\linewidth]{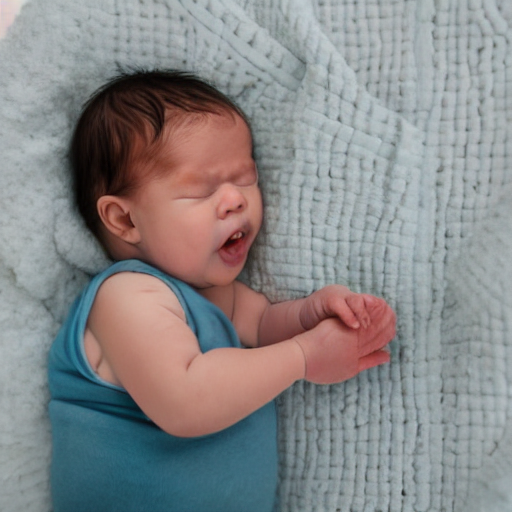}}
	\end{minipage}                                                                & \begin{minipage}[b]{0.25\columnwidth}
		\raisebox{-.5\height}{\includegraphics[width=\linewidth]{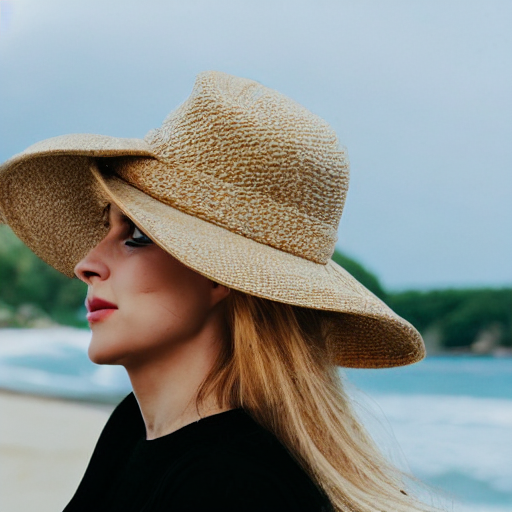}}
	\end{minipage}                                                                      & \begin{minipage}[b]{0.25\columnwidth}
		\raisebox{-.5\height}{\includegraphics[width=\linewidth]{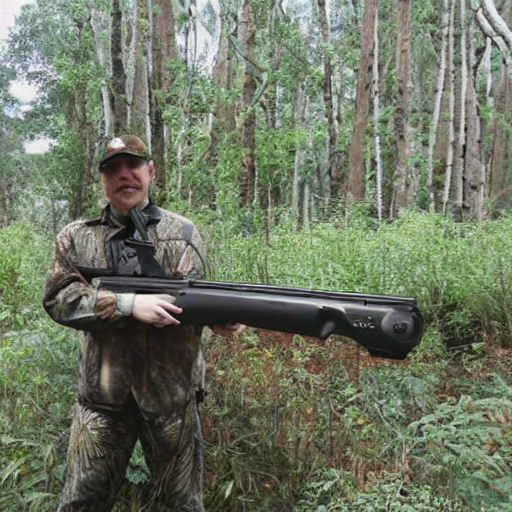}}
	\end{minipage}                                                                     & \begin{minipage}[b]{0.25\columnwidth}
		\raisebox{-.5\height}{\includegraphics[width=\linewidth]{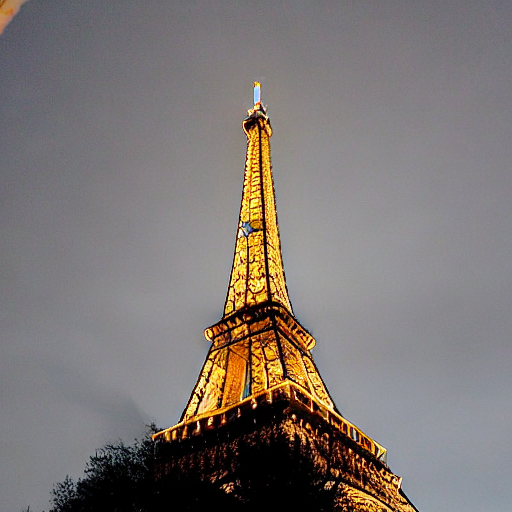}}
	\end{minipage}                                                                    &    \begin{minipage}[b]{0.25\columnwidth}
		\raisebox{-.5\height}{\includegraphics[width=\linewidth]{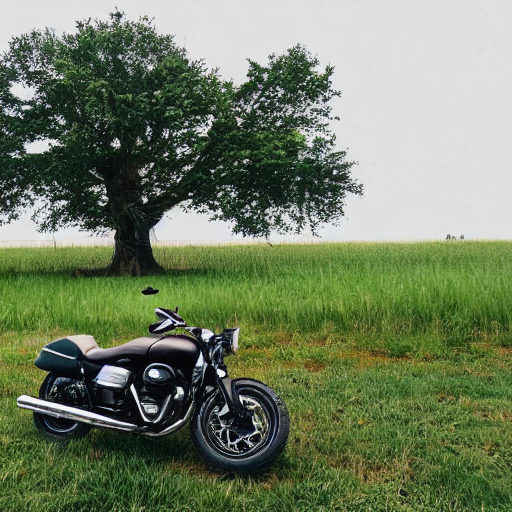}}
	\end{minipage}                    
                       \\ \hline \hline
    \multicolumn{5}{c}{Prompt: His guitar by the couch waits for practice}                       \\ \hline \hline
     \begin{minipage}[b]{0.25\columnwidth}
		\raisebox{-.5\height}{\includegraphics[width=\linewidth]{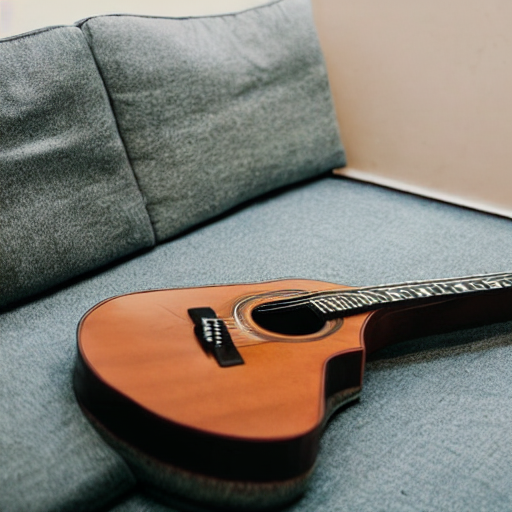}}
	\end{minipage}                                                                & \begin{minipage}[b]{0.25\columnwidth}
		\raisebox{-.5\height}{\includegraphics[width=\linewidth]{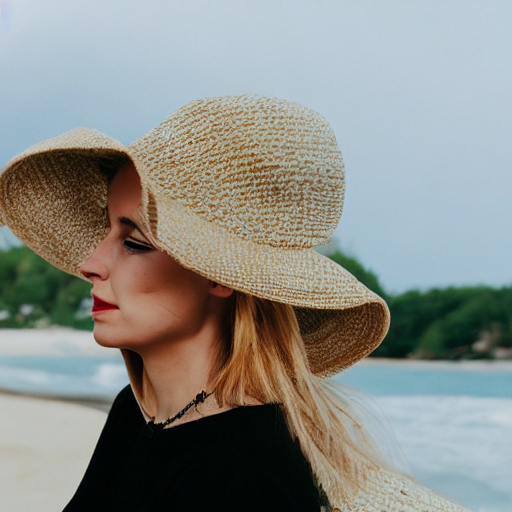}}
	\end{minipage}                                                                      & \begin{minipage}[b]{0.25\columnwidth}
		\raisebox{-.5\height}{\includegraphics[width=\linewidth]{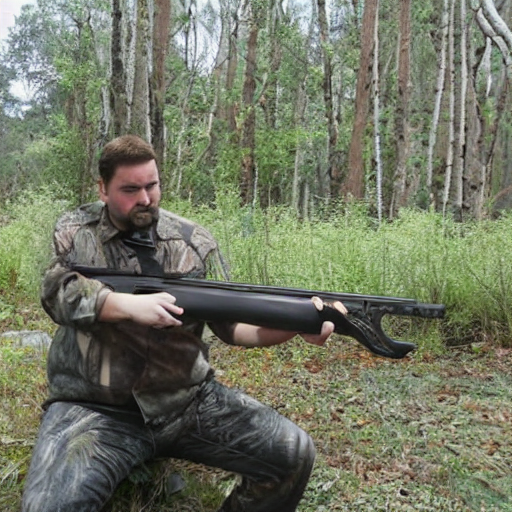}}
	\end{minipage}                                                                     & \begin{minipage}[b]{0.25\columnwidth}
		\raisebox{-.5\height}{\includegraphics[width=\linewidth]{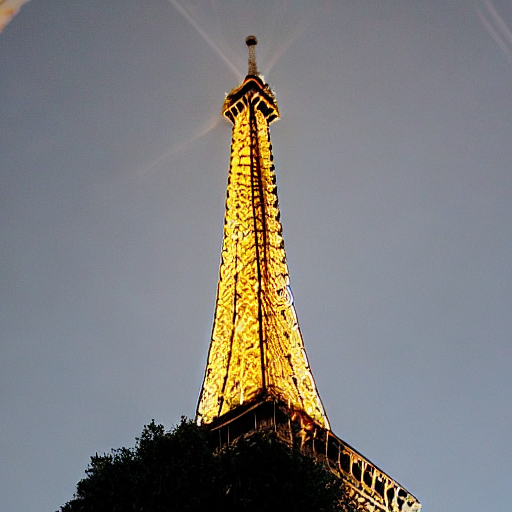}}
	\end{minipage}                                                                    &    \begin{minipage}[b]{0.25\columnwidth}
		\raisebox{-.5\height}{\includegraphics[width=\linewidth]{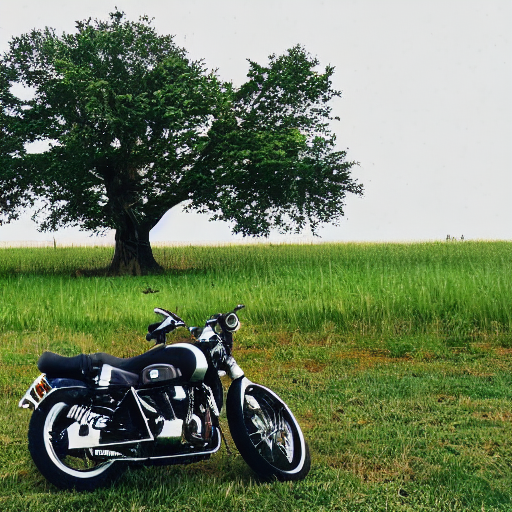}}
	\end{minipage}                    
                       \\ \hline \hline
    \multicolumn{5}{c}{Prompt: These flowers in the vase bloom with beauty}                       \\ \hline \hline
     \begin{minipage}[b]{0.25\columnwidth}
		\raisebox{-.5\height}{\includegraphics[width=\linewidth]{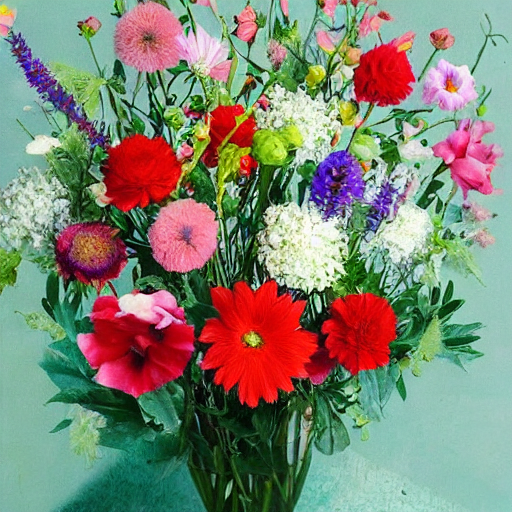}}
	\end{minipage}                                                                & \begin{minipage}[b]{0.25\columnwidth}
		\raisebox{-.5\height}{\includegraphics[width=\linewidth]{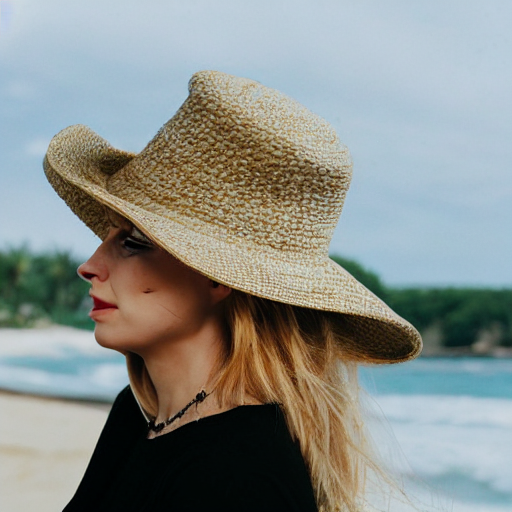}}
	\end{minipage}                                                                      & \begin{minipage}[b]{0.25\columnwidth}
		\raisebox{-.5\height}{\includegraphics[width=\linewidth]{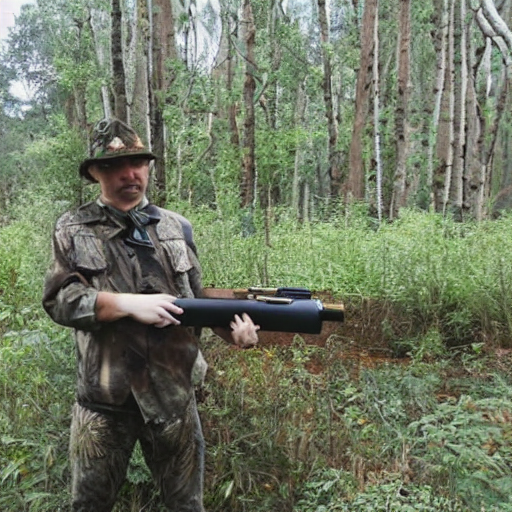}}
	\end{minipage}                                                                     & \begin{minipage}[b]{0.25\columnwidth}
		\raisebox{-.5\height}{\includegraphics[width=\linewidth]{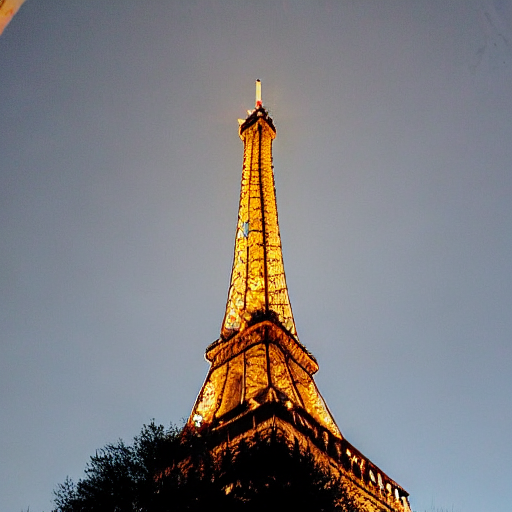}}
	\end{minipage}                                                                    &    \begin{minipage}[b]{0.25\columnwidth}
		\raisebox{-.5\height}{\includegraphics[width=\linewidth]{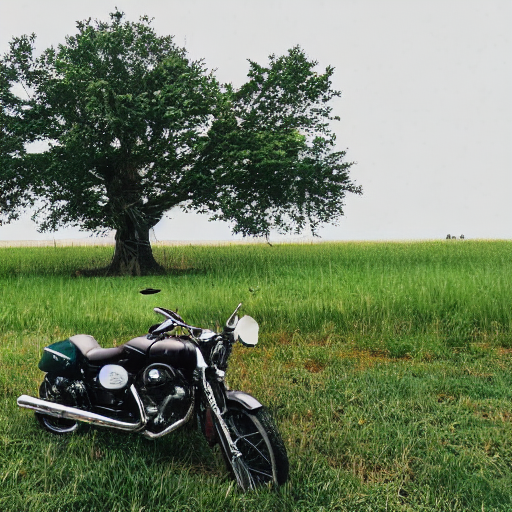}}
	\end{minipage} \\ \hline \hline
    \multicolumn{5}{c}{Prompt: Her bike in the garage leans near door}                       \\ \hline \hline
         \begin{minipage}[b]{0.25\columnwidth}
		\raisebox{-.5\height}{\includegraphics[width=\linewidth]{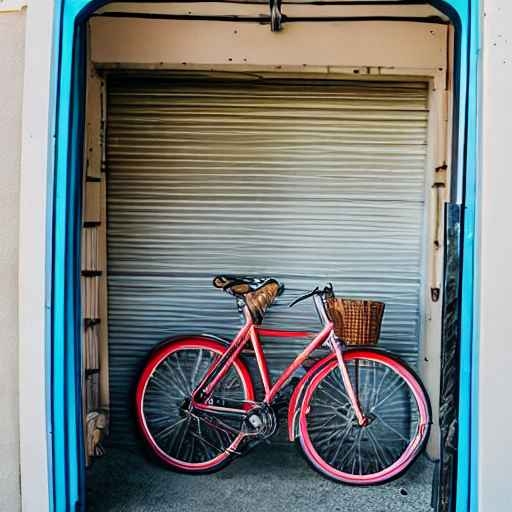}}
	\end{minipage}                                                                & \begin{minipage}[b]{0.25\columnwidth}
		\raisebox{-.5\height}{\includegraphics[width=\linewidth]{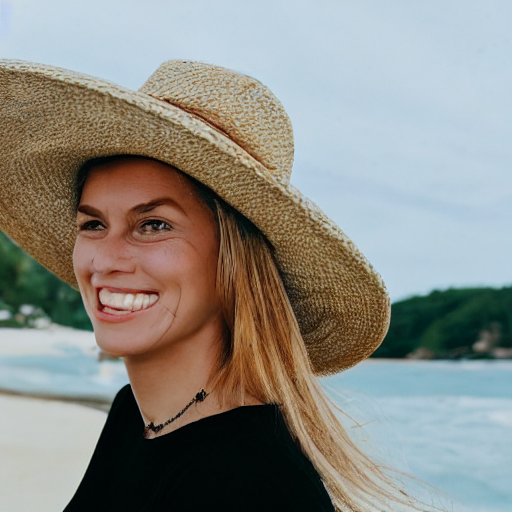}}
	\end{minipage}                                                                      & \begin{minipage}[b]{0.25\columnwidth}
		\raisebox{-.5\height}{\includegraphics[width=\linewidth]{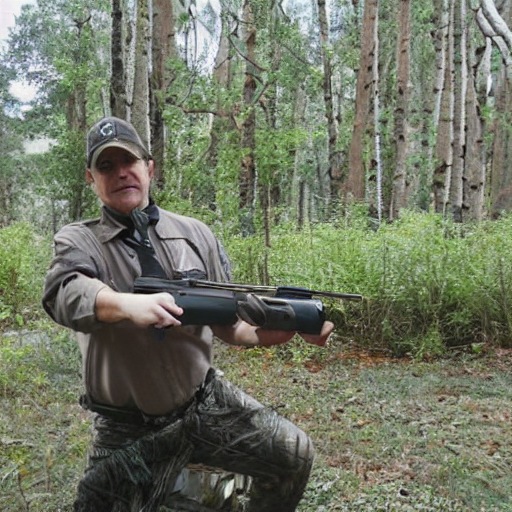}}
	\end{minipage}                                                                     & \begin{minipage}[b]{0.25\columnwidth}
		\raisebox{-.5\height}{\includegraphics[width=\linewidth]{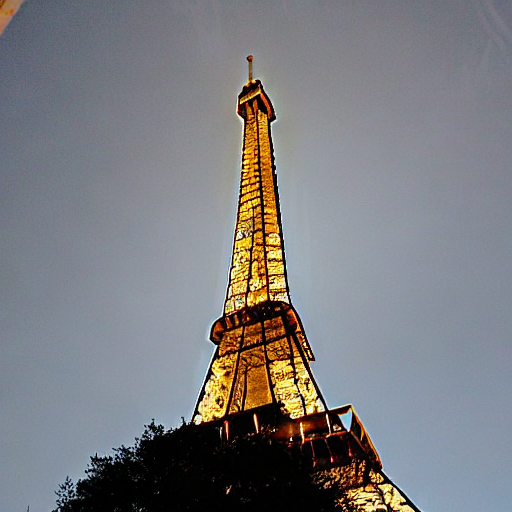}}
	\end{minipage}                                                                    &    \begin{minipage}[b]{0.25\columnwidth}
		\raisebox{-.5\height}{\includegraphics[width=\linewidth]{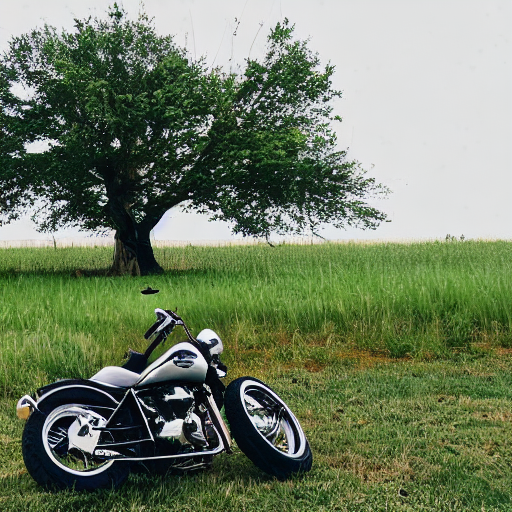}}
	\end{minipage}                    
    \\  \hline
\end{tabular}

\label{tab:adr}
\end{table*}

\vfill

\end{document}